%% file: main.tex
\documentclass[10pt]{article} 
\usepackage[preprint]{tmlr}

\usepackage{hyperref} 
\usepackage{url}
\input{support}
\usepackage{cleveref}
\usepackage[nohints]{minitoc}

\title{Sublinear Algorithms for Estimating Wasserstein and TV Distances: Applications to Fairness and Privacy Auditing}

\author{\name Debabrota Basu \email debabrota.basu@inria.fr\\
      \addr \'Equipe Scool, Univ. Lille, Inria,\\ 
  CNRS, Centrale Lille, UMR 9189- CRIStAL\\ 
  F-59000 Lille, France\\
      \ANDAUTH
      \name Debarshi Chanda \email debarshi.chanda.1997@gmail.com\\
      \addr Indian Statistical Institute\\
      Kolkata, India
      }

\usepackage[dvipsnames]{xcolor}

\hypersetup{colorlinks,citecolor=Cerulean,linkcolor=Purple}

\def\month{01}  
\def\year{2026} 
\def\openreview{}

\begin{document}

\maketitle

\doparttoc
\faketableofcontents

\begin{abstract}
Resource-efficiently computing representations of probability distributions and the distances between them while only having access to the samples is a fundamental and useful problem across mathematical sciences. In this paper, we propose a generic framework to learn the probability and cumulative distribution functions (PDFs and CDFs) of a sub-Weibull, i.e. almost any light- or heavy-tailed, distribution while the samples from it arrive in a stream. The idea is to reduce these problems into estimating the frequency of an \textit{appropriately chosen subset} of the support of a \textit{properly discretised distribution}. We leverage this reduction to compute mergeable summaries of distributions from the stream of samples while requiring only sublinear space relative to the number of observed samples. This allows us to estimate Wasserstein and Total Variation (TV) distances between any two distributions while samples arrive in streams and from multiple sources. Our algorithms significantly improves on the existing methods for distance estimation incurring super-linear time and linear space complexities, and further extend the mergeable summaries framework to continuous distributions with possibly infinite support. Our results are tight with respect to the existing lower bounds for bounded discrete distributions. In addition, we leverage our proposed estimators of Wasserstein and TV distances to tightly audit the fairness and privacy of algorithms. We empirically demonstrate the efficiency of proposed algorithms across synthetic and real-world datasets.
\end{abstract}

\thispagestyle{empty}
\newpage


\setcounter{page}{1}

\input{Sections_PODS/0_Introduction}

\input{Sections_PODS/1_Prelims}

\input{Sections_PODS/2_Problem_Setup}
\input{Sections_PODS/3_Technical}

\input{Sections_PODS/4_ApplicationAuditing}

\input{Sections_PODS/6_Conclusion}
\section*{Acknowledgement}
Both the authors would like to thank the Inria-ISI Kolkata associate team SeRAI (DRI-012551) for supporting this collaboration. D.~Basu acknowledges ANR JCJC project REPUBLIC (ANR-22-CE23-0003-01) and PEPR project FOUNDRY (ANR23-PEIA-0003) partially funding this work. We also thank Uddalok Sarkar and Udvas Das for their support.

\clearpage
\bibliography{icml2024/refs_ICML}
\bibliographystyle{tmlr}



\newpage
\onecolumn
\appendix
\part{Appendix}
\vfill
\parttoc
\vfill

\input{Sections_PODS/Appendix}

\end{document}

%% file: support.tex
\newif\iffinal
\finaltrue

\usepackage{amsmath,amsfonts, mathtools}

\usepackage{algorithm, algorithmic}

\usepackage{amsthm, thmtools, amssymb}

\theoremstyle{plain}
\newtheorem{theorem}{Theorem}
\newtheorem{lemma}[theorem]{Lemma}
\newtheorem{corollary}[theorem]{Corollary}

\newtheorem{assumption}[theorem]{Assumption}

\theoremstyle{definition}
\newtheorem{definition}[theorem]{Definition}

\theoremstyle{remark}
\newtheorem{remark}[theorem]{Remark}

\input{Support/boxes}
\input{Support/fonts}

\input{Support/macros}

\newcommand{\field}[1]{\mathbb{#1}}

\newcommand{\R}{\field{R}}
\newcommand{\Nat}{\field{N}}

\newcommand{\dd}{\mathrm{d}}

\newcommand{\norm}[1]{\left\|{#1}\right\|}

\newcommand{\func}[3]{{#1} : {#2} \rightarrow {#3}}
\DeclarePairedDelimiter{\ceil}{\lceil}{\rceil}

\newcommand{\wh}{\widehat}

\newcommand{\blue}[1]{{#1}}

\newcommand{\sequence}[2]{{#1}_1,{#1}_2,...,{#1}_{#2}}
\newcommand{\fbrac}[1]{\left({#1}\right)}
\newcommand{\sbrac}[1]{\left\{{#1}\right\}}
\newcommand{\tbrac}[1]{\left[{#1}\right]}
\newcommand{\abs}[1]{\left|{#1}\right|}
\newcommand{\size}[1]{\left|{#1}\right|}
\newcommand{\constant}{c}

\newcommand{\dom}{\mathop{\mathrm{Dom}}}
\newcommand{\range}{\mathop{\mathrm{Range}}}
\newcommand{\lipschitzconstant}{\ell}
\newcommand{\upperlipc}{\lipschitzconstant}
\newcommand{\lowerlipc}{\frac{1}{\lipschitzconstant}}
\newcommand{\lipschitz}{Lipschitz }
\newcommand{\indexset}{I}

\newcommand{\Wasserstein}[3]{\operatorname{\sW_{#1}}\left({#2},{#3}\right)}
\newcommand{\TV}[2]{\operatorname{TV}\left({#1},{#2}\right)}
\newcommand{\hockeystick}[3]{\operatorname{H_{#1}}\fbrac{{#2}||{#3}}}

\newcommand{\hamming}[2]{\operatorname{Ham}\fbrac{{#1},{#2}}}

\DeclareMathOperator*{\Prob}{\field{P}}
\DeclareMathOperator*{\Exp}{\field{E}}

\newcommand{\support}[1]{\mathrm{Supp}\fbrac{#1}}
\newcommand{\distribution}{\sD}
\newcommand{\uniform}{\mathrm{U}}
\newcommand{\normal}{\sN}

\newcommand{\fhat}{\hat{f}}
\newcommand{\invfhat}{\hat{f}^{-1}}
\newcommand{\approxerror}{\varepsilon}
\newcommand{\confidence}{\delta}
\newcommand{\inverse}[1]{{#1}^{-1}}

\newcommand{\bigo}[1]{\mathcal{O}\fbrac{{#1}}}
\newcommand{\bigot}[1]{\widetilde{\mathcal{O}}\fbrac{{#1}}}
\newcommand{\smallo}[1]{o\fbrac{{#1}}}

\newcommand{\bigomega}[1]{\Omega\fbrac{{#1}}}

\newcommand{\merge}{\texttt{Merge}}



%% file: Support/boxes.tex
\usepackage[most]{tcolorbox}
\newtcolorbox{idea}[1][]
{
colbacktitle=cyan,
colback=cyan!10,
arc=1pt,
boxrule=1pt,
title=#1 
}

\newtcolorbox{update}[1][]
{
colbacktitle=gray,
colback=gray!10,
arc=1pt,
boxrule=1pt,
title=#1 
}

\newtcolorbox{question}[1][]
{
coltitle=black,
colbacktitle=yellow,
colback=yellow!10,
arc=1pt,
boxrule=1pt,
title=#1 
}

\newtcolorbox{note}[1][]
{
coltitle=black,
colbacktitle=green,
colback=green!10,
arc=1pt,
boxrule=1pt,
title=#1 
}

\newtcolorbox{answer}[1][]
{
coltitle=black,
colbacktitle=violet!10,
colback=violet!5,
arc=1pt,
boxrule=1pt,
title=#1 
}

%% file: Support/fonts.tex
\usepackage{bm}

\newcommand{\bw}{\bm{w}}

\newcommand{\sD}{\mathcal{D}}

\newcommand{\sF}{\mathcal{F}}

\newcommand{\sM}{\mathcal{M}}
\newcommand{\sN}{\mathcal{N}}

\newcommand{\sP}{\mathcal{P}}

\newcommand{\sU}{\mathcal{U}}

\newcommand{\sW}{\mathcal{W}}
\newcommand{\sX}{\mathcal{X}}
\newcommand{\sY}{\mathcal{Y}}


%% file: Support/macros.tex
\newcommand{\stream}{\zeta}
\newcommand{\streamelement}[1]{\stream_{#1}}
\newcommand{\streamlength}{n}
\newcommand{\streamweight}[1]{\bw_{#1}}
\newcommand{\streamuniverse}{\sU}
\newcommand{\universeelement}[1]{\streamuniverse_{#1}}
\newcommand{\streamuniversesize}{m}
\newcommand{\truefrequency}[1]{f_{#1}}
\newcommand{\truecumulant}[1]{F_{#1}}
\newcommand{\estfreqset}{\widehat{\sF}}
\newcommand{\estimatedfrequency}[1]{\hat{f}_{#1}}
\newcommand{\estimatedcumulant}[1]{\hat{F}_{#1}}
\newcommand{\totalfrequency}{\wassmeassamplecount}
\newcommand{\residualfrequency}[1]{\blue{{\mathrm{res}(\totalfrequency,{#1})}}}
\newcommand{\countersize}{k}
\newcommand{\counterthreshold}{\countersize^*}

\newcommand{\measure}{\mu}
\newcommand{\weibull}{\alpha}
\newcommand{\threshold}{\tau}
\newcommand{\clients}{S}

\newcommand{\metricspace}{\sX}
\newcommand{\wassmeas}{\mu}
\newcommand{\wassmeastwo}{\nu}
\newcommand{\estwassmeas}{{\hat{\wassmeas}_\wassmeassamplecount}}
\newcommand{\estwassmeastwo}{\hat{\wassmeastwo}_\wassmeassamplecount}

\newcommand{\bucketelement}{\sqcup}
\newcommand{\bucketsize}{b}
\newcommand{\bucketset}{\mathcal{B}}

\newcommand{\bucketwassmeas}[1]{\wassmeas_{#1}}

\newcommand{\bucketdisc}[2]{{#1}_{#2}^{Disc}}
\newcommand{\bucketcont}[2]{{#1}_{#2}^{Cont}}

\newcommand{\wassmeassamplecount}{n}
\newcommand{\empwassmeas}{\wassmeas_\wassmeassamplecount}
\newcommand{\empwassmeastwo}{\wassmeastwo_\wassmeassamplecount}
\newcommand{\empbucketwassmeas}[1]{\wassmeas_{{#1},\wassmeassamplecount}}
\newcommand{\empbucketwassmeastwo}[1]{\wassmeastwo_{{#1},\wassmeassamplecount}}

\newcommand{\subgaussianparameter}{\sigma}

\newcommand{\tail}{t}
\newcommand{\CDF}{Q}
\newcommand{\pdf}{p}
\newcommand{\estCDF}{\widehat{\CDF}}
\newcommand{\estpdf}{\hat{\pdf}}
\newcommand{\iid}{i.i.d. }
\newcommand{\invCDF}{\CDF^{-1}}
\newcommand{\estinvCDF}{\widehat{\invCDF}}
\newcommand{\couplingset}[2]{\Pi\left({#1},{#2}\right)}
\newcommand{\coupling}{\pi}

\newcommand{\dkwerror}{\approxerror_{\scriptscriptstyle{DKW}}}

\newcommand{\topk}[1]{Top_{#1}}

\newcommand{\mergablemisragries}{\texttt{MMG}}
\newcommand{\slwassbase}{\texttt{SWA}}
\newcommand{\slCDFbase}{\texttt{SCA}}
\newcommand{\slPDFbase}{\texttt{SPA}}
\newcommand{\slTVbase}{\texttt{STVA}}

\newcommand{\baseapprox}{\approxerror}

\newcommand{\initializestream}{\texttt{Initialize}}
\newcommand{\updatestream}{\texttt{Update}}
\newcommand{\estimatestream}{\texttt{Estimate}}
\newcommand{\estimatecumulant}{\texttt{EstimateCumulate}}

\newcommand{\basebucketwidth}{\ceil*{\frac{8\subgaussianparameter_\wassmeas}{\bucketsize}\sqrt{\log{\fbrac{\frac{4}{\baseapprox}}}}}}

\newcommand{\basemgcover}{\ceil*{2\subgaussianparameter_\wassmeas\sqrt{\log{\fbrac{\frac{4}{\baseapprox}}}}}}

\newcommand{\basemglengthpdf}{\ceil*{\frac{8\subgaussianparameter_\wassmeas}{\bucketsize}\sqrt{\log{\fbrac{\frac{6}{\baseapprox}}}}}}

\newcommand{\basemgcoverpdf}{\ceil*{2\subgaussianparameter_\wassmeas\sqrt{\log{\fbrac{\frac{6}{\baseapprox}}}}}}

\newcommand{\dpepsilon}{\alpha}
\newcommand{\dpdelta}{\beta}
\newcommand{\dppair}{$(\dpepsilon,\dpdelta)$}

\newcommand{\covariatematrix}{\mathbf{\ \MakeUppercase{\covariatei}}}
\newcommand{\covariate}{\mathbf{\covariatei}}
\newcommand{\covariatei}{x}

\newcommand{\covariatedomain}{\sX}
\newcommand{\targetrange}{\sY}
\newcommand{\mechanism}{\sM}

\newcommand{\acsincome}{ACS\textunderscore Income }

\newcommand{\nthres}[1]{\threshold_\streamlength\fbrac{#1}}
\newcommand{\kthres}[1]{\threshold_\countersize\fbrac{#1}}
\newcommand{\bthres}[1]{\threshold_\bucketsize\fbrac{#1}}

\newcommand{\countersum}[1]{C_{#1}}
\newcommand{\frequencysum}[1]{N_{#1}}
\newcommand{\errorsum}[1]{E_{#1}}
\newcommand{\counter}[1]{c_{#1}}
\newcommand{\decrement}{\texttt{DecrementCounters}}
\newcommand{\deccounter}{\counter{\counterthreshold}}

%% file: Sections_PODS/0_Introduction.tex
\section{Introduction}

Computing distances between probability distributions while having access to samples is a fundamental and practical problem in statistics, computer science, information theory, and many other fields. 
Among the multitude of distances proposed between distributions, we focus on the Wasserstein and Total Variation distances in this work. 

\textbf{Wasserstein Distance.} Optimal transport~\citep{Villani2009} has emerged as a popular and useful tool in machine learning. It involves computing Wasserstein distance between distributions and using it further for learning. The statistical and computational aspects of Wasserstein distances have been extensively studied, refer to~\citet{Panaretos2018StatisticalAO,Book/Computational_OT/Cuturi,chewi2024statisticaloptimaltransport}. This geometrically insightful approach has found applications in text document similarity measurement~\citep{kusner2015word}, dataset distances~\citep{alvarez2020geometric}, domain adaption~\citep{courty2016optimal}, generative adversarial networks~\citep{arjovsky2017wasserstein}, data selection and valuation~\citep{justlava,kang2024performance}, fair classification and regression~\citep{pmlr-v115-jiang20a,DBLP:conf/nips/ChzhenDHOP20} to name a few. Thus, it is a fundamental question to compute Wasserstein distance between two distributions when we only have access to samples from it. But computing Wasserstein distance is often memory and computationally intensive until we know the exact parametric form of the data distribution (precisely, location-scatter families)~\citep{GelbrichWasserSteinCharecterization,ALVAREZESTEBAN2016744,CuestaAlbertosOneDWasserstein,alvarez2020geometric}. In absence of such parametric assumptions, we need linear space complexity and super-linear time complexity in terms of observed samples to estimate Wasserstein distances between the underlying data distributions~\citep{Cuturi/Sinkhorn/Neurips13,Panaretos2018StatisticalAO,Chizat/Sinkhorn/Neurips20,DBLP:conf/iclr/RakotomamonjyNR24,chewi2024statisticaloptimaltransport}. 
Additionally, in practice, the stream might arrive from multiple sources, as in Federated Learning~\citep{kairouz2021advances}. This has motivated a recent line of work to estimate Wasserstein distance in federated setting~\citep{DBLP:conf/iclr/RakotomamonjyNR24,li2024privatewassersteindistancerandom}. However, these methods incur communication cost equal to the number of samples in the stream. 
These gaps in literature motivate us to ask
\begin{center}
    \blue{Can we estimate the Wasserstein distance between two continuous distributions with infinite support\\ in sublinear space, time, and communication complexity w.r.t. number of samples $n$?}
\end{center}

\textbf{Total Variation (TV) Distance.} TV~\citep{Devroye/AOS/1990/EmpircalTVDoesntConverge} is another well-studied distance between two probability distributions. It measures the maximum gap between the probabilities of any event w.r.t. the two distributions. TV also quantifies the minimum probability that $X\neq Y$ among all couplings of $(X, Y)$ sampled from the distributions. TV distance between the output distributions of an algorithm operated on two neighbouring datasets also central to auditing the privacy level of the algorithm~\citep{koskela2024auditingdifferentialprivacyguarantees}. Additionally, in bandits, the hardness of a problem depends on the TV distance between the reward distributions of the available actions~\citep{azize2022,azize2023}. Thus, estimating TV distance from empirical samples emerges as a fundamental problem in privacy and machine learning. 
Estimating TV distance has been studied extensively~\citep{ClementCanonne/TopicsTechniquesDistributionTesting}. There is a long line of work focusing on estimating TV distance between discrete distributions with finite support~\citep{pmlr-v40-Kamath15/Colt/DiscreteDistributionTV,Han/ISIT/2015/MinimaxEstimationofDiscreteDistributions,Feng/SODA/2024/DeterministicApproximationTV, Devroye/EJS/2019/DiscreteMinimaxTVwithTrees, Bhattacharyya/IJCAI/2023/OnApproximatingTotalVariationDistance, bhattacharyya/arxiv/2024/computationalexplorationstotalvariation,  Bhattacharyya/ICML/2024/TVmeetsProbabilisticInference}. \citet{Feigenbaum/SIAMJComp/2002/ApproxL1Streaming,GuhaMcGregor/SODA/2006/DivergencesDiscreateStreaming,RoyVasudev/ISAAC/2023/TestingDistributionPropertiesStreaming} propose sublinear algorithm to estimate TV distance while having access to sample streams. But the question of estimating TV distance for continuous distributions with infinite supports remains wide open. This motivates the question:
\begin{center}
    \textit{Can we estimate the TV distance between two continuous distributions with infinite support\\ in sublinear space, time, and communication complexity w.r.t. number of samples $n$?}
\end{center}

We address these questions for any bounded tail distribution by \textit{reducing them into estimation of a sublinear summary, i.e. the frequency of an appropriately chosen subset of the support of a properly discretised distribution}.

\textbf{Mergeable Sublinear Representation of Distributions.} We propose to learn an approximately correct sublinear summary of a distribution from the data stream that can address the previous questions (Section~\ref{sec:reduction}). Because then we can store, communicate, and compute with only these sublinear and discrete summaries of distributions. In literature, we find that the sublinear summaries of histograms has been extensively studied over decades~\citep{MISRA1982143,AMSSketch,CountSketch,CountMinSketch}. Additionally, in the continuous distributed monitoring setup~\citep{cormode2013continuous}, one might want to estimate TV and Wasserstein distances when data streams arrive from multiple sources. Also, with the growth of federated and distributed learning, the question of \textit{how these sublinear-sized summaries of histograms from multiple streams can be merged efficiently} has gained interest~\citep{Mergeable_Summaries, Berinde_Mergeable,Efficient_Mergable_Misra_Gries}. 
In machine learning, we work a lot with continuous data distributions. Thus, to be practically useful, we need to extend these sublinear summarisers to continuous setting and find minimal conditions to yield theoretical guarantees. On the other hand, there is another long line of research to learn histograms from samples of a continuous distribution and control the error in this process~\citep{ScottHistogramBins1979,FreedmanDiaconis/ZWVG/1981/HistogramBinWidthForDensityEstimation,ioannidis2003history,diakonikolas2018fast}. But it is an open question to develop algorithms and analysis to join these two streams of research, i.e.
\begin{center}
    \textit{Can we learn sublinear summaries of (possibly continuous) distributions\\ while having a stream of $n$ samples from it?}
\end{center}

\textbf{Our contributions} address these questions affirmatively. 

1. \textit{Computing Mergeable Sublinear Summary from Multiple Streams.} We propose a generic framework to learn summaries of PDF and CDF of a continuous or discrete (possible with infinite support) distribution using sublinear space (Section~\ref{sec:learnpdfcdf}). Given data arriving in a stream (or multiple streams) from an underlying distribution, we propose \slPDFbase{} and \slCDFbase{} $\epsilon$-approximate corresponding PDF and CDF independent of stream length. To our knowledge, \textit{we initiate the study of {mergeable sublinear summaries} over streams from an infinite set} and \textit{establish theoretical error bounds for streams from an infinite set and any sub-Gaussian or sub-Weibull distribution}.

2. \textit{Sublinearly Estimating Wasserstein distance.} We use \slCDFbase{} to propose \slwassbase{} to PAC-estimate Wasserstein distance (Section~\ref{sec:estimateW}). We first show that \textit{mergeable} sublinear summaries learned by \slCDFbase{} are universal estimators of the true distribution in Wasserstein distance. In turn, \slwassbase{} computes a sublinear summary of a distribution that is $\bigo{n^{-1/2}}$ close in Wasserstein distance using $\bigot{\sqrt{n}}$ space for $n$ samples\footnote{Soft-O notations, i.e. $\bigot{\cdot}$ and $\widetilde{\Theta}\fbrac{\cdot}$, ignore the polylogarithmic and other lower order terms.} \slwassbase{} operates in the distributed/federated setting with a communication cost of $\bigot{\sqrt{n}}$ per round while preserving the estimation guarantee.

3. \textit{Sublinearly Estimating TV distance.} For TV distance, we leverage \slPDFbase{} and set the parameters properly to propose \slTVbase{}  (Section~\ref{sec:estimateTV}). \slTVbase{} maintains mergeable sublinear summaries of the bucketed versions of the true distributions, which have probably infinite buckets in their supports. We show that if we set the bucket width to $\widetilde{\Theta}(n^{-1/3})$, \slTVbase{} yields a $\bigot{n^{-1/3}}$ estimate of TV distance between two distributions using $\bigot{n^{1/3}}$ space.

4. \textit{Applying estimators for fairness and privacy auditing.} In Section~\ref{sec:audits}, we demonstrate usefulness of our Wasserstein and TV distances estimators for auditing fairness and privacy of machine learning models, respectively. Experimental results demonstrate the accuracy and sublinearity of the proposed estimators for auditing models trained on real-life datasets from fairness and privacy literatures.

%% file: Sections_PODS/1_Prelims.tex
\section{Preliminaries}
We discuss the fundamentals of distributional distances and the generic algorithmic template to learn mergeable summaries of histograms, which are essential to this work. 

\textbf{Notations.} $\wassmeas$, $\empwassmeas$ and $\empbucketwassmeas{\bucketsize}$ denote the true, empirical, and bucketed empirical measure with bucket width $\bucketsize$, respectively.  $\func{\pdf_\wassmeas}{\support{\wassmeas}}{\tbrac{0,1}}$ and $\func{\CDF_\wassmeas}{\support{\wassmeas}}{\tbrac{0,1}}$ denote the probability density function (PDF) and cumulative density function (CDF) of $\wassmeas$, respectively. $\tbrac{n}$ refers to $\{1,2,\ldots,n\}$ for $n\in\mathbb{N}$.

\subsection{Distances between Probability Distributions}

Now, we define the Wasserstein and TV distances between distributions. We start with defining the Wasserstein distance between distributions defined over metric spaces.

\begin{definition}[p-th Wasserstein distance~\citep{Villani2009}]\label{Definition: Wasserstein Distance}
Given two probability measures $\wassmeas,\wassmeastwo$ over a metric space $\metricspace$, the p-th Wasserstein distance between them is
\begin{align*}
\Wasserstein{p}{\wassmeas}{\wassmeastwo} \triangleq \left(\min_{\coupling \in \couplingset{\wassmeas}{\wassmeastwo}} \int_{\metricspace\times\metricspace} \norm{x - y}^p \,\dd\coupling(x,y) \right)^\frac{1}{p}\,.
\end{align*}
$\pi$ is a coupling over $x$ and $y$, and $\Pi$ is the set of all couplings.
\end{definition}\vspace*{-.5em}

Here, Wasserstein distance is the cost calculated for the optimal coupling between the two probability measures over Euclidean distances. For univariate distributions, Wasserstein distance can be expressed as norms between inverse CDFs of the distributions~\citep{Book/Computational_OT/Cuturi}.

\begin{lemma}\label{Lemma: Univariate Wasserstein Characterization}
For univariate measures $\wassmeas,\wassmeastwo$ over $\R$, the $p$-th Wasserstein distance $\Wasserstein{p}{\wassmeas}{\wassmeastwo}$ is
\begin{align}
\Wasserstein{p}{\wassmeas}{\wassmeastwo} 
= \norm{\invCDF_\wassmeas - \invCDF_\wassmeastwo}_{L^p([0,1])}^p \,,\label{Equation: Univariate Wasserstein Distance}
\end{align}
where $\func{\invCDF_\wassmeas}{\tbrac{0,1}}{\R\cup\sbrac{-\infty}}$ of a probability measure $\wassmeas$ is the pseudoinverse function defined as $
    \invCDF_\wassmeas(r) \triangleq \min_{x \in \R \cup \sbrac{-\infty}} \sbrac{x : \CDF_\wassmeas(x) \geq r}$ for $r \in [0,1]$.
\end{lemma}

The other distance that we study is the Total Variation (TV) distance.

\begin{definition}[Total Variation (TV) Distance]\label{Definition: TV Distance}
    Given measures $\wassmeas,\wassmeastwo$ with support $\mathcal{X}$, TV distance between them is \vspace*{-.8em}
    \begin{align*}
        \TV{\wassmeas}{\wassmeastwo} \triangleq \sup_{A\subseteq \mathcal{X}}|\wassmeas(A) - \wassmeastwo(A)|= \frac{1}{2}\norm{\wassmeas-\wassmeastwo}_1.
    \end{align*}
\end{definition}

%


\subsection{Mergable Summaries of Histograms}
The principal algorithmic technique that we use is that of summarising a histogram, which is a long-studied problem~\citep{MISRA1982143,AMSSketch,CountSketch,Mergeable_Summaries,Efficient_Mergable_Misra_Gries}. We refer to~\citet{Cormode_Survey_Heavy_Hitters} for an overview.   

First, we formally introduce the problem. We consider data arriving in a stream $\stream$ of length $\streamlength$, where each element belongs to a universe $\streamuniverse$. The $j$-th element of $\streamuniverse$, i.e. $\universeelement{j}$, appears $\truefrequency{j}$ times in the stream $\stream$. Our goal is to maintain an approximate count of elements $\estimatedfrequency{j}$ such that $\norm{\estimatedfrequency{j}-\truefrequency{j}}_{p}$ is small for some $p$. If we separately keep count of all the elements , the problem is trivial and we get $\estimatedfrequency{j} = \truefrequency{j}, \forall \universeelement{j} \in \streamuniverse$. However, the goal is to obtain a `good' approximation while using sublinear, i.e. $\smallo{n}$, space. 
More generally, the $i$-th element of the stream can have a weight $\streamweight{i} \in \Nat$. Then, the task is to generate estimates $\estfreqset = \sbrac{\estimatedfrequency{j}| \universeelement{j} \in \streamuniverse}$, which are close to the true frequency $\truefrequency{j} = \sum_{i=1}^n \streamweight{i}\mathbf{1}[\streamelement{i} = \universeelement{j}]$ for all $j$. 

Now, we extend the problem setup to aggregate from multiple streams~\citep{Mergeable_Summaries}. Given $\clients$-streams of data $\stream_1, \stream_2,...,\stream_\clients$, we aim to generate estimates $\estfreqset_1, \estfreqset_2,...,\estfreqset_\clients$ and \textit{combine them efficiently} to output a globally `good' frequency estimate $\estfreqset = \texttt{merge}\fbrac{\estfreqset_1,\estfreqset_2,...,\estfreqset_\clients}$ for the concatenated stream $\stream_1 \circ \stream_2 \circ ... \circ \stream_\clients$.

We build on the algorithm of~\citet{Efficient_Mergable_Misra_Gries}, which extends the well-known Misra-Gries algorithm~\citep{MISRA1982143}.  We refer to it as the Mergeable Misra-Gries (\mergablemisragries{}) algorithm and provide the details in Algorithm~\ref{Algo: Mergable Misra-Gries}. This family of algorithms is broadly known as the \textit{counter-based algorithms for histogram summarisation} and consists of \textit{three main modules}. First, they operate by maintaining a sublinear number of counters, say $\countersize = o(n)$, all initiated with $0$. Then, the module $\mergablemisragries{}\cdot\updatestream{}$ updates them according to the elements in stream. Specifically, for each stream element, if our algorithm has already assigned a counter, it adds the weight to the corresponding counter. If not, it assigns an empty counter to the element if there is an unassigned counter. Otherwise, the values of all counters are reduced by the median value of counters. Second, if the streams are coming from multiple sources, the module $\mergablemisragries{}\cdot\merge{}$ takes the counters in all the summaries as an input stream, and update each counter in the final summary by considering the counters as stream elements and their counts as the corresponding weights. For updating, it reuses the module $\mergablemisragries{}\cdot\updatestream{}$. 
Finally, given an element, the module $\mergablemisragries{}\cdot\estimatestream{}$ returns an estimate of its frequency by returning the value stored in the counter if the element is assigned a counter, and $0$ otherwise. Note that \updatestream{}, \estimatestream{}, and \merge{} takes amortised $\bigot{1},\bigot{1}$, and $\bigot{\countersize}$ time to execute, respectively. For brevity, we defer the further details to Appendix~\ref{Appendix: MMG Algo Exposition}. 

Now, we derive a rectified version of Theorem 2 of~\citet{Efficient_Mergable_Misra_Gries} that bounds the error in frequency estimation using the residuals $\residualfrequency{\countersize/4}$ (defined below). 


\begin{restatable}[Estimation Guarantee of \mergablemisragries]{lemma}{MMGGuarantee}\label{Theorem: Mergeable Misra Gries}
    (a) If \mergablemisragries{} uses $\countersize$ counters, $\mergablemisragries\cdot\estimatestream$ yields a summary $\{\estimatedfrequency{j}\}_{\universeelement{j}\in\streamuniverse}$ satisfying $ 0 \leq \truefrequency{j} - \estimatedfrequency{j} \leq \frac{\residualfrequency{\tau}}{\fbrac{\countersize-\counterthreshold} - \tau}$, 
    for all $\tau \leq \fbrac{\countersize-\counterthreshold}$ and $\universeelement{j} \in\abs{\streamuniverse}$. Here, $\residualfrequency{\tau}$ denotes the sum of frequency of all but $\tau$ most frequent items. 
    (b) If we choose $\counterthreshold = \frac{\countersize}{2}$ $\tau = \countersize/4$, we get that $0 \leq \truefrequency{j} - \estimatedfrequency{j} \leq \frac{4~\residualfrequency{\countersize/4}}{\countersize}$, $\forall \universeelement{j} \in\abs{\streamuniverse}$.
\end{restatable}

We \textit{extend and analyse \mergablemisragries's design technique, originally developed for the discrete distributions with finite support points, to continuous and discrete distributions with infinite support points}.


%% file: Sections_PODS/2_Problem_Setup.tex
\section{Problem: Estimating Distances between Distributions from Sample Streams}
Now, we formally state our problem setup. 
Let $\wassmeas$ and $\wassmeastwo$ be two measures on $\R$. 
We consider that the data is arriving in two streams $\stream_{\wassmeas}$ and $\stream_{\wassmeastwo}$ of size $\wassmeassamplecount$, and each element of the two streams are independent and identically distributed (i.i.d.) samples of $\wassmeas$ and $\wassmeastwo$, respectively. We denote by $\estwassmeas$ and $\estwassmeastwo$ the two summaries of $\wassmeas$ and $\wassmeastwo$ computed from the sample streams. 
Given a probability metric $\mathfrak{D}$, our objective is to compute an $(\approxerror,\confidence)$-estimate of $\mathfrak{D}\fbrac{\wassmeas,\wassmeastwo}$ while using sublinear ($\smallo{\wassmeassamplecount}$) space to store $\estwassmeas$ and $\estwassmeastwo$.
Specifically, we want to yield $\mathfrak{D}\fbrac{\estwassmeas,\estwassmeastwo}$ ensuring
\begin{align}
    \Prob\tbrac{| \mathfrak{D}\fbrac{\wassmeas,\wassmeastwo} - \mathfrak{D}\fbrac{\estwassmeas,\estwassmeastwo} | \geq \approxerror} \leq \confidence\,,
\end{align}
such that $|\mathrm{essSup}(\estwassmeas) \cup \mathrm{essSup}(\estwassmeastwo)|=\smallo{\wassmeassamplecount}$, and $(\approxerror, \confidence) \in (0,1) \times (0,1)$. 
In this work, we particularly focus on two probability metrics: Wasserstein distances $\mathfrak{D}\fbrac{\wassmeas,\wassmeastwo} = \Wasserstein{p}{\wassmeas}{\wassmeastwo}$, and TV distance $\mathfrak{D}\fbrac{\wassmeas,\wassmeastwo} = \TV{\wassmeas}{\wassmeastwo}$. 
For streams coming from multiple sources, like in the {federated} (or continuous distributed monitoring) setting, we consider the concatenated streams $\stream_{\wassmeas} = \stream_{\wassmeas,1} \circ \stream_{\wassmeas,2} \circ \ldots \circ \stream_{\wassmeas,\clients}$ and $\stream_{\wassmeastwo} = \stream_{\wassmeastwo,1} \circ \stream_{\wassmeastwo,2} \circ \ldots \circ \stream_{\wassmeastwo,\clients}$, to be the input streams of length $n$ each to estimate the distance $\mathfrak{D}\fbrac{\wassmeas,\wassmeastwo}$.

\paragraph{Structural Assumptions.} In this work, we assume that the data generating distributions have bounded tails (Assumption~\ref{Assumption: Bounded Tails}). For the case of $\mathfrak{D}\fbrac{\wassmeas,\wassmeastwo} = \Wasserstein{p}{\wassmeas}{\wassmeastwo}$, we assume the distribution to have $\lipschitzconstant_\distribution$ bi-Lipschitz CDF (Assumption~\ref{Assumption: Bi-Lipschitz Distributions}). For the case of $\mathfrak{D}\fbrac{\wassmeas,\wassmeastwo} = \TV{\wassmeas}{\wassmeastwo}$, we assume the distribution to have $\lipschitzconstant_\distribution$-Lipschitz PDF (Assumption~\ref{Assumption: Lipschitz PDF}). 

To formalise the bounded-tail assumption, we first define the sub-Gaussian and sub-Weibull random variables.
\begin{definition}[$\fbrac{\threshold,\subgaussianparameter}$-Sub-Gaussian Distributions]\label{Def: Subgaussianity}
    A distribution $\wassmeas$ is said to be $\fbrac{\threshold,\subgaussianparameter}$-sub-Gaussian if for any $\threshold \geq \tail \geq 0$, we have $\Prob_{X\sim \distribution}\left[|X - \Exp[X]| \geq \tail\right] \leq 2\exp{\left(-\frac{\tail^2}{\subgaussianparameter_\wassmeas^2}\right)}\,$.
\end{definition}
When $\threshold \rightarrow \infty$, $\fbrac{\threshold,\subgaussianparameter}$-sub-Gaussian distribution reduces to the classical definition of sub-Gaussian distribution~\citep{Vershynin_2018}. Sub-Gaussians cover a wide-range of distributions including any bounded distribution, Gaussians, mixture of Gaussians etc.~\citep{Vershynin_2018}. It is standard to assume the noise and data to be sub-Gaussian in regression problems~\citep{wainwright2019high}. Sub-Gaussianity appears in the underlying scoring mechanism for classification~\citep{wang2018optimal}.
We also consider the notion of sub-Weibull distribution that generalises the notion of sub-Gaussianity to any form of exponentially bounded tails. While sub-Gaussian distributions are considered to be light-tailed, sub-Weibull distributions are considered for heavy-tail distributions in the concentration of measure literature~\citep{foss2011introduction,VladimirovaGirardNguyenArbel/Stat/2020/SubWeibullDistributions,bakhshizadeh2023sharp}.

\begin{definition}[$\fbrac{\threshold,\weibull}$-Sub-Weibull Distributions]\label{Def: Partial sub-Weibull}
    A distribution $\wassmeas$ is said to be $\fbrac{\threshold,\weibull}$-sub-Weibull if there exists some constant $\constant_\weibull$ for any $\threshold \geq \tail \geq 0$, we have $\Pr_{X\sim\wassmeas}\tbrac{X \geq \tail} \leq \constant_\weibull\exp\fbrac{-\tail^{1/\weibull}}\,.$.
\end{definition}
   When $\threshold \rightarrow \infty$, $\fbrac{\threshold,\weibull}$-sub-Weibull distribution reduces to the classical definition of sub-Weibull distribution~\citep{VladimirovaGirardNguyenArbel/Stat/2020/SubWeibullDistributions}. For simplicity, we denote $\fbrac{\infty,\weibull}$-sub-Weibull Distributions as ${\weibull}$-sub-Weibull. Note that, for sub-Gaussians, $\weibull = 1/2$. Now, we formally state our assumptions. 


\begin{assumption}[Sub-Gaussian or sub-Weibull Data Generating Distributions]\label{Assumption: Bounded Tails}
    We consider the distributions yielding the samples are either $\subgaussianparameter$-sub-Gaussian or $\weibull$-sub-Weibull.
\end{assumption}

The tail bound assumption is required because we have access to only the samples from the stream rather than the true distribution. Thus, we need to have \textit{enough} samples such that the empirical measures available to the algorithm are \textit{close} to the true distributions. The tail bounds provide a control of this concentration of measure phenomenon. \blue{We now state our assumptions regarding the lipschitzness of the distributions.}
\blue{\begin{definition}[$\ell$-Lipschitz Functions]\label{Def: Lipschitzness}
    Given two metric spaces $(\mathbf{X},d_x)$ and $(\mathbf{Y},d_y)$, a function $f:\mathbf{X}\rightarrow\mathbf{Y}$ is said to be $\ell$-Lipschitz if $d_y(f(x_1),f(x_2)) \leq \ell d_x(x_1,x_2) ~~ \forall x_1,x_2 \in \mathbf{X}$.
\end{definition}}

\begin{assumption}[Bi-Lipschitz Distributions]\label{Assumption: Bi-Lipschitz Distributions}
The distribution $\distribution$ over $(\mathbf{X},d_x)$ with CDF $\CDF_\distribution$ has $\lipschitzconstant_\distribution$-bi-Lipschitz CDF, if $\frac{1}{\lipschitzconstant_\distribution} d_x(a,b) \leq \abs{\CDF_\distribution(a) - \CDF_\distribution(b)} \leq \lipschitzconstant_\distribution d_x(a,b)\,,~\forall a,b \in \mathbf{X}\,.$
\end{assumption}
This essentially implies that the PDF of $\distribution$ is bounded above and below everywhere. \blue{Any distribution with bounded PDF, and} continuous and compact support satisfies this, which encompasses most of the common data distributions.

\begin{assumption}[Lipschitz PDF Distributions]\label{Assumption: Lipschitz PDF}
    The distribution $\distribution$ with pdf $\measure_\distribution$ over $(\mathbf{X},d_x)$ has $\lipschitzconstant_\distribution$-Lipschitz PDF, i.e. $
    	\abs{\measure_\distribution(a) - \measure_\distribution(b)} \leq \lipschitzconstant_\distribution d_x(a,b)\,,~\forall a,b \in \mathbf{X}$.
\end{assumption}
Any exponential family distribution with bounded parameters have Lipschitz PDF. It includes Gaussians with bounded mean and variance, exponential and Poisson distributions, and other commonly used distributions.

%% file: Sections_PODS/3_Technical.tex
\section{Sublinear Estimators of Wasserstein and TV Distances}

In this section, we explain our methodology for mergeable summary creation of bounded-tail distributions, and estimation of Wasserstein and TV distances from sample streams. We theoretically and numerically demonstrate accuracy and sublinearity of our estimators. 

\subsection{From Sublinear Distance Estimation to Learning Sublinear Summaries}\label{sec:reduction}
Let us consider the empirical measure $\empwassmeas$ on $\wassmeassamplecount$ points generated from the true (possibly continuous) measure $\wassmeas$.  
Thus, for any probability distance $\mathfrak{D}$, we observe that
\begin{align}\label{equation:error_decompose}
       &| \mathfrak{D}\fbrac{\wassmeas,\wassmeastwo} - \mathfrak{D}\fbrac{\estwassmeas,\estwassmeastwo} | \leq \underbrace{\mathfrak{D}\fbrac{\wassmeas,\empwassmeas} + \mathfrak{D}\fbrac{\wassmeastwo,\empwassmeastwo}}_{\substack{\text{Concentration of measures}}} + \underbrace{\mathfrak{D}\fbrac{\empwassmeas,\estwassmeas} +\mathfrak{D}\fbrac{\empwassmeastwo,\estwassmeastwo}}_{\substack{\text{Sublinear summaries}}}\,,\notag
\end{align}
if all of these distances are well-defined. 
In that case, the distance of the empirical measure and the true one decreases due to concentration of measures. While we learn sublinear summaries for each of the empirical distributions and aim to control the distance between the empirical distributions and the sublinear summary, i.e. $\mathfrak{D}\fbrac{\empwassmeas,\estwassmeas}$, for any  bounded-tail true distribution $\wassmeas$. 

In Section~\ref{sec:learnpdfcdf}, we propose algorithms to estimate PDFs and CDFs of bounded-tail distributions. This allows us to control $\mathfrak{D}\fbrac{\empwassmeas,\estwassmeas}$ and in turn, compute the desired distances if $\empwassmeas$ has bounded-tails. We first proof that the empirical distribution $\empwassmeas$ corresponding to a sub-Gaussian or sub-Weibull $\wassmeas$ is also sub-Gaussian or sub-Weibull, respectively.


%
\blue{
\begin{restatable}[Empirical Measure is Sub-Gaussian]{theorem}{EmpircalMeasureSubGaussian}\label{Theorem: Empirical Measure is Subgaussian}
    Given an $\subgaussianparameter$-sub-Gaussian true measure $\wassmeas$, the empirical measure $\empwassmeas$ generated by $\sequence{X}{\streamlength}$ drawn i.i.d. from $\wassmeas$ is $\fbrac{\threshold,\subgaussianparameter}$-sub-Gaussian with probability at least $1 - \confidence$ given $\streamlength \geq \constant\exp\fbrac{\frac{\threshold^2}{\subgaussianparameter^2}}\log\frac{1}{\confidence}$.
\end{restatable}}

\begin{restatable}[Empirical Distribution is sub-Weibull]{theorem}{EmpircalMeasureSubWeibull}\label{theorem: Empirical Distribution is SubWeibull}
    Given an $\weibull$-sub-Weibull true measure $\wassmeas$, the empirical measure $\empwassmeas$ generated by $\sequence{X}{\streamlength}$ drawn i.i.d. from $\wassmeas$ is $\fbrac{\threshold,\weibull}$-sub-Weibull with probability at least $1 - \confidence$ given $\streamlength \geq \frac{\exp\fbrac{\sqrt[\weibull]{\threshold}}}{12}\log\frac{1}{\confidence}$.
\end{restatable}
We show that with high probability, the empirical distribution retains sub-Gaussianity and sub-Weibull property of the true measure $\wassmeas$. This result is important for learning sublinear summaries of true distribution from sample streams. The proof is in Appendix~\ref{app:empiricaldist}.


\subsection{Learning Mergable Sublinear Summaries}\label{sec:learnpdfcdf}
In this section, we state our results regarding sublinear summary approximation of sub-Gaussian and sub-Weibull distributions satisfying Assumption~\ref{Assumption: Bounded Tails}. \blue{We use \mergablemisragries{} (\Cref{Algo: Mergable Misra-Gries}) that approximates a data stream generated from bounded number of elements. We design appropriate bucketing techniques, and leverage the tail bounds to approximate the PDF and CDF of continuous distributions using this algorithm.}



\noindent\textbf{A. Learning PDF.} We first turn the support of the empirical distribution into a collection of buckets of width $b>0$, and in turn, construct an empirical bucketed distribution $\empbucketwassmeas{\bucketsize}$ out of the empirical distribution $\empwassmeas$.

\begin{definition}[\textbf{Bucketed Empirical Distribution}]\label{Definition: Bucketed Empirical Distribution}
  Let $\distribution$ be a distribution supported on a (finite or infinite) closed interval $\bucketset \subseteq \R$ with (discrete or continuous) measure $\measure$. Given a reference point $x_0 \in \bucketset$, bucket width $\bucketsize$, and an index set $\indexset \subseteq \mathbb{Z}$, we can represent $\bucketset = \bucketset(x_0, \indexset, \bucketsize) \triangleq \cup_{i \in \indexset} [x_0 + i\bucketsize, x_0 + (i+1)\bucketsize]$. We define the bucketed empirical distribution, denoted $\empbucketwassmeas{\bucketsize}$, to be the empirical distribution generated from $\wassmeassamplecount$ samples of {$\wassmeas$} as $    \empbucketwassmeas{\bucketsize}(i) \triangleq \frac{\wassmeassamplecount_i}{\wassmeassamplecount}$, where $\wassmeassamplecount_i$ is the number of samples falling in the bucket $\bucketelement_i$.
\end{definition}
For brevity, we denote the bucket $[x_0 + i\bucketsize, x_0 + (i+1)\bucketsize]$ by $\bucketelement_i$. We note that $\bucketset$ inherits the metric structure of $\R$. Specifically, the distance between two points $x_1 \in \bucketelement_i$ and $x_2 \in \bucketelement_j$ is defined as $d(x_1, x_2) = |i-j|b$. 



In Algorithm~\ref{Algorithm: TV Producer}, we first fix a number of buckets $k$, and treat each of the buckets as an element of the stream. The weight of a bucket is the number of elements in the stream that falls in it, which is $n_i$ for $\bucketelement_i$. 
We apply \mergablemisragries{} over these buckets to return the estimated frequency $\hat{f}_i$ for each bucket $\bucketelement_i$. Finally, we return the learned PDF $\{\pdf_{\empbucketwassmeas{\bucketsize}}(i)\}_{i\in\indexset} = \Big\lbrace \frac{\estimatedfrequency{i}}{\max_{j}\estimatedcumulant{j}}\Big\rbrace_{i\in\indexset}$ as a summary of $\empwassmeas$'s PDF.
Now, we bound the learning error of \slPDFbase. 

\begin{minipage}{0.48\textwidth}
\begin{algorithm}[H]
    \caption{Sublinear PDF Approximator:\\ ${\slPDFbase}(\stream \overset{\iid}{\gets} \wassmeas,\countersize,\bucketsize)$}\label{Algorithm: TV Producer}
    \begin{algorithmic}[1]
        \STATE \textbf{\initializestream}{}
        \STATE $\mergablemisragries{}\cdot\initializestream\fbrac{\countersize}$
        \STATE \textbf{\updatestream}{$\fbrac{\streamelement{i}}$}
            \STATE $\bucketelement_{\streamelement{i}} \gets \bucketelement$ that contains $\streamelement{i}$
            \STATE $\mergablemisragries{}\cdot\updatestream\fbrac{\streamelement{i},1}$

        \STATE \textbf{\merge(}$T_2$\textbf{)}
            \STATE $\mergablemisragries{}\cdot\merge\fbrac{T_2}$
        \STATE \textbf{\estimatestream}{$\fbrac{\streamelement{i}}$}
            \STATE $\estimatedfrequency{i} \gets \mergablemisragries{}\cdot\estimatestream{\fbrac{\streamelement{i}}}$
            \STATE return $\pdf_{\estwassmeas}(i) \gets \frac{\estimatedfrequency{i}}{\max_{j}\estimatedcumulant{j}}$ 
    \end{algorithmic}
\end{algorithm}

\end{minipage}\hfill
\begin{minipage}{0.48\textwidth}

\begin{algorithm}[H]
    \caption{Sublinear CDF Approximator \\${\slCDFbase}(\stream \overset{\iid}{\gets} \wassmeas,\approxerror,\countersize,\bucketsize)$}\label{Algorithm: CDF Producer}
    \begin{algorithmic}[1]
        \STATE \textbf{\initializestream}{}
        \STATE $\mergablemisragries{} \cdot \initializestream\fbrac{\countersize}$
        
        \STATE \textbf{\updatestream}{$\fbrac{\streamelement{i}}$}
            \STATE $\bucketelement_{\streamelement{i}} \gets \bucketelement$ that contains $\streamelement{i}$
            \STATE $\mergablemisragries{}\cdot\updatestream\fbrac{\bucketelement_{\streamelement{i}},1}$

        \STATE \textbf{\merge(}$T_2$\textbf{)}
            \STATE $\mergablemisragries{}\cdot\merge\fbrac{T_2}$
        
        \STATE \textbf{\estimatestream}{$\fbrac{\streamelement{i}}$}
            \STATE $\estimatedcumulant{j} \gets \mergablemisragries{}\cdot\estimatecumulant{\fbrac{\streamelement{i}}}$
            \STATE return $\estCDF_{i} \gets \frac{\estimatedcumulant{i}}{\max_{j}\estimatedcumulant{j}}$              
    \end{algorithmic}
\end{algorithm}
\end{minipage}\vskip 1em

\begin{restatable}[\slPDFbase{} Learning Error]{theorem}{slPDFbaseWorks}\label{Theorem: slPDFbase works}
    (a) If \slPDFbase{} uses $\countersize$ buckets and outputs $\estwassmeas$, then for all $i \in \indexset$, 
    \begin{align*}      
        \abs{\pdf_{\estwassmeas}(i) - \pdf_{\empbucketwassmeas{\bucketsize}}(i)} &\leq \max\sbrac{\frac{4~\residualfrequency{\countersize/4}}{\totalfrequency}\frac{\truefrequency{i}}{\totalfrequency},\frac{\truefrequency{i} - \estimatedfrequency{i}}{\totalfrequency}}\,.
    \end{align*}
    (b) Further, if $\wassmeas$ corresponding to $\empwassmeas$ is $\subgaussianparameter_\wassmeas$-sub-Gaussian, \blue{$\size{\stream} = \wassmeassamplecount \geq \frac{\constant}{\approxerror}\log\fbrac{\frac{1}{\confidence}}$}, and $\countersize \geq \ceil*{\frac{8\subgaussianparameter_\wassmeas}{\bucketsize}\sqrt{\log{\fbrac{\frac{1}{\baseapprox}}}}}$, then for all $i \in \indexset$, $\mathbb{P}(\abs{\pdf_{\estwassmeas}(i) - \pdf_{\empbucketwassmeas{\bucketsize}}(i)} \geq \approxerror) \leq \delta.$
    
(c) Further, if $\wassmeas$ corresponding to $\empwassmeas$ is $\weibull$-SubWeibull, $\size{\stream} = \wassmeassamplecount \geq \frac{\constant}{\approxerror}\log\fbrac{\frac{1}{\confidence}}$, and $\countersize \geq \ceil*{\frac{\constant}{\bucketsize}\fbrac{\log\fbrac{\frac{1}{\approxerror}}}^\weibull}$, then for all $i \in \indexset$, $\mathbb{P}(\abs{\pdf_{\estwassmeas}(i) - \pdf_{\empbucketwassmeas{\bucketsize}}(i)} \geq \approxerror) \leq \delta.$
\end{restatable}
Lemma~\ref{Theorem: Mergeable Misra Gries} shows that the error in \mergablemisragries{} depends on the sum of frequency of elements except the ones with the highest frequencies. \slPDFbase{} leverages this observation, the boundedness of the tails of a distribution (Assumption~\ref{Assumption: Bounded Tails}), and the guarantee on the preservation of tail bounds for the empirical measure (Theorem~\ref{Theorem: Empirical Measure is Subgaussian}~and~\ref{theorem: Empirical Distribution is SubWeibull}). In \slPDFbase, we set bucket width $\bucketsize$ as $\countersize \geq \kthres{\approxerror,\bucketsize}$ such that the (possibly) uncovered tails have sufficiently small mass. This yields an $(\epsilon,\delta)$-PAC estimate of the empirical PDF. {The proof is in Appendix~\ref{Appendix: Performance Guarantee of SPA and SCA}.}
\input{icml2024/Figures/summary_figures}

\noindent\textbf{B. Learning CDF.} Now, we want to learn a sublinear summary of the CDF of the empirical distribution $\empwassmeas$. Thus, we start by bucketing its support with $k$ counters. Given the corresponding PDF $\pdf_{\empbucketwassmeas{\bucketsize}}$, we define the CDF of the bucketed empirical distribution as $\CDF_{\empbucketwassmeas{\bucketsize}}(i) \triangleq \sum_{j\leq i} \pdf_{\empbucketwassmeas{\bucketsize}}(j)$. 

In general, if there is a \textit{total order} on the stream universe $\streamuniverse$, we can define the cumulant function $\func{\truecumulant{}}{\streamuniverse}{\Nat}$ as $\truecumulant{i} \triangleq \sum_{j \leq i} \truefrequency{j}$. Total order exists for the real line and the bucket $\bucketset$ defined on it.  Lemma~\ref{Lemma: Estimate Cumulant Works} establishes the error bound of $\mergablemisragries{}\cdot\estimatecumulant{}$ for the cumulant $\truecumulant{}$. {The proof is in Appendix~\ref{Appendix: MMG Algo Exposition}.}

\begin{restatable}[Error Bound of $\mergablemisragries{}\cdot\estimatecumulant$]{lemma}{cumulateestimateguarantee}\label{Lemma: Estimate Cumulant Works}
    If \mergablemisragries{} uses $\countersize$ counters, $\mergablemisragries\cdot\estimatecumulant{}$ yields a summary $\{\estimatedcumulant{j}{}\}_{\universeelement{j}\in\streamuniverse}$ satisfying $0 \leq \truecumulant{i} - \estimatedcumulant{i} \leq 2\residualfrequency{\frac{\countersize}{4}}\,.$\vspace*{-.5em}
\end{restatable}

Now, in the spirit of Theorem~\ref{Theorem: slPDFbase works}, we appropriately set the bucket width and collect enough samples to yield an $(\epsilon,\delta)$-PAC estimate of the empirical CDF (Appendix~\ref{Appendix: Performance Guarantee of SPA and SCA}). 

\begin{restatable}[\slCDFbase{} Learning Error]{theorem}{slCDFbaseWorks}\label{Theorem: slCDFbase works}
    Given stream length $\size{\stream} = \wassmeassamplecount \geq \nthres{\approxerror,\confidence} $  generated from a bounded-tail distribution $\wassmeas$, \slCDFbase{} with bucket width $\bucketsize$ sets \#buckets to $\countersize \geq \kthres{\approxerror,\bucketsize} $ and outputs $\estCDF{}$, such that with probability at least $1 - \confidence$, $\abs{\CDF_{\empbucketwassmeas{\bucketsize}}(i) - \estCDF(i)} \leq \approxerror\,$ for all $i \in \indexset$.
    \renewcommand{\arraystretch}{1.6}
    \begin{table}[H]
        \centering
        \begin{tabular}{c|cc}
        Tail Condition    &  $\nthres{\approxerror,\confidence}$ &  $\kthres{\approxerror,\bucketsize}$\\
        \hline
        $\subgaussianparameter_\wassmeas$-sub-Gaussian   &  \blue{$\frac{\constant}{\approxerror}\log\fbrac{\frac{1}{\confidence}}$} &  $\basebucketwidth$ \\
        $\weibull$-sub-Weibull    &  $\frac{\constant}{\approxerror}\log\fbrac{\frac{1}{\confidence}}$   &   $\ceil*{\frac{\constant}{2\bucketsize}\fbrac{\log\fbrac{\frac{4}{\approxerror}}}^\weibull}$\\
        \hline
        \end{tabular}
        \label{tab:slCDFbaseBounds}
    \end{table}
\end{restatable}
Note that the space complexities of both \slPDFbase{} and \slCDFbase{} are independent of streaming universe's size.

\textbf{Numerical Demonstration.} We take a stream of $10^5$ samples from a Gaussian $\mathcal{N}(0,5)$ and estimate its PDF with \slPDFbase. We set the bucket width to $\bucketsize=0.05$ and vary the number of buckets from $100$ to $700$. Figure~\ref{Fig: pdf learn} validates that increasing the number of buckets, while still being sublinear in \#samples, in \slPDFbase{} leads to more accurate summaries of a continuous sub-Gaussian density function.


\subsection{Estimating Wasserstein Distance}\label{sec:estimateW}

In this section, we introduce an algorithm, namely \slwassbase{}, to estimate Wasserstein distance between distributions satisfying Assumption~\ref{Assumption: Bounded Tails} and~\ref{Assumption: Bi-Lipschitz Distributions}. \slwassbase{} leverages \slCDFbase{} to learn the CDF and use the inverse CDF formulation of Wasserstein (Lemma~\ref{Lemma: Univariate Wasserstein Characterization}) to obtain a good estimation of the Wasserstein distance between two distributions. We present the pseudocode of \slwassbase{} in Algorithm~\ref{Algorithm: Base Wasserstein Approximation Algorithm}.

\begin{algorithm}[ht!]
    \caption{{Sublinear Wasserstein Estimator}: ${\slwassbase}(\stream_{\wassmeas}, \stream_{\wassmeastwo} \overset{\iid}{\gets} \wassmeas, \wassmeastwo, \approxerror, p)$}\label{Algorithm: Base Wasserstein Approximation Algorithm}
    \begin{algorithmic}[1]
        \REQUIRE $\lipschitzconstant_\wassmeas, \lipschitzconstant_\wassmeastwo \gets$ Lipschitz constant of $\wassmeas$ and $\wassmeastwo$, $p \gets$ $p$-th Wasserstein
        \FOR{$\distribution \in \{\wassmeas, \wassmeastwo\}$}
            \STATE $\bucketsize_{\distribution} \gets \frac{\epsilon}{4}$
            \STATE $\{\CDF_{\hat{\distribution}_n}(i)\}_{i=1}^k \gets \slCDFbase{\fbrac{\stream_{\distribution},\bucketsize_{\distribution},\kthres{\approxerror,\lipschitzconstant_\distribution,\confidence},\approxerror/2}}$\label{SWA Line: SCA Call}
        \ENDFOR
            \STATE \textbf{return} $\Wasserstein{p}{\estwassmeas}{\estwassmeastwo}\gets$ plug $\{\CDF_{\estwassmeas}(i)\}_{i=1}^k$ and $\{\CDF_{\estwassmeastwo}(i)\}_{i=1}^k$ in Equation~\eqref{Equation: Univariate Wasserstein Distance}
    \end{algorithmic}
\end{algorithm}

\textbf{Universal Learning Guarantee of \slCDFbase.} We show that \slCDFbase{} is a universally good estimator of the CDF of any bounded-tail measure with respect to the Wasserstein distance. \textit{This is a culmination of three results.}

\underline{First}, \slCDFbase{} guarantees an $(\epsilon,\delta)$-PAC summary of the CDF of a bucketed empirical distribution. To leverage that guarantee in the case of Wasserstein distance for a (possibly continuous) measure $\wassmeas$, we need to choose the bucket width $\bucketsize$ appropriately such that $\Wasserstein{p}{\empwassmeas}{\empbucketwassmeas{\bucketsize}}$ is small as the distance of the true and the bucketed distribution is bounded by $b$ (Lemma~\ref{Lemma: Bucketing Wasserstein Approximation}). 

\underline{Second}, once we have the $(\epsilon,\delta)$-PAC summary of the CDF, the rest is to show that it yields small error in computing the inverse CDF, and in turn, the Wasserstein distance (Equation~\eqref{Equation: Univariate Wasserstein Distance}). We establish a guarantee on the approximation error of the pseudoinverse of a bi-Lipschitz CDF, given a guarantee on the approximation error on the CDF (Corollary~\ref{Corollary: Inverse CDF Approximation}).


\underline{Finally}, we show that the empirical measure $\empwassmeas$ corresponding to a true measure $\wassmeas$ with bi-Lipschitz CDF also exhibits bi-Lipschitz CDF with high probability. 
\begin{restatable}[{Bucketed} Empirical Measure is Bi-Lipschitz]{theorem}{EmpiricalMeasureBiLipschitz}\label{Theorem: Empirical Measure is Bi-Lipschitz}
    Let the bucketed empirical measure $\empbucketwassmeas{\bucketsize}$ generated by $\wassmeassamplecount$ i.i.d. samples coming from a distribution $\wassmeas$ with $\lipschitzconstant_{\wassmeas}$ bi-\lipschitz CDF. For any $\approxerror, \confidence \in (0,1)$ and fixed constant $c>0$, if $\wassmeassamplecount \geq \frac{\constant}{\approxerror^2\bucketsize^2}\log\fbrac{\frac{2}{\confidence}}\max\sbrac{\frac{1}{\lipschitzconstant_{\wassmeas}^2},\lipschitzconstant_{\wassmeas}^2}$, $\empbucketwassmeas{\bucketsize}$ is $(1+\approxerror)\lipschitzconstant_{\wassmeas}$ bi-\lipschitz with probability $1-\confidence$. 
\end{restatable}

{Proof of Theorem~\ref{Theorem: Empirical Measure is Bi-Lipschitz} is in Appendix~\ref{app:empiricaldist}}. Finally, these results together yield Theorem~\ref{Theorem: slWassBase Works}, i.e. \slCDFbase{} learns a universal estimator of any measure $\wassmeas$. {The proof is in Appendix~\ref{Appendix: SWA Guarantees}.}

\begin{restatable}[Universal Learning of \slCDFbase{} in $\Wasserstein{p}{}{}$]{theorem}{slWassBaseWorks}\label{Theorem: slWassBase Works}
    Given $\size{\stream} = \wassmeassamplecount \geq \nthres{\lipschitzconstant_\wassmeas,\approxerror,\confidence} $ generated from a bounded-tail distribution $\wassmeas$ and $\approxerror, \confidence \in (0,1)$, if we set the bucket width $\bucketsize=\approxerror/2$, then \slCDFbase{} uses $\kthres{\lipschitzconstant_\wassmeas,\approxerror} $ buckets (i.e. space) and outputs $\estwassmeas$, such that $\mathbb{P}(\Wasserstein{p}{\empwassmeas}{\estwassmeas} \geq \approxerror)\leq \confidence$. We omit the constants in $\nthres{\lipschitzconstant_\wassmeas,\approxerror,\confidence}$ and $\kthres{\lipschitzconstant_\wassmeas,\approxerror}$ for simplicity.
    \begin{table}[H]
        \centering
        \begin{tabular}{c|cc}
        Tail Condition    &  $\nthres{\lipschitzconstant_\wassmeas,\approxerror,\confidence}$ &  $\kthres{\lipschitzconstant_\wassmeas,\approxerror}$\\
        \hline
        $\subgaussianparameter_\wassmeas$-sub-Gaussian   &  \blue{$\log\fbrac{1/\confidence}\max\sbrac{\frac{1}{\approxerror},\frac{1}{\lipschitzconstant_{\wassmeas}^2},\lipschitzconstant_{\wassmeas}^2}$} &  $\frac{\subgaussianparameter_\wassmeas}{\approxerror}\log\fbrac{\frac{\lipschitzconstant_\wassmeas}{\approxerror}}$ \\
        $\weibull$-sub-Weibull    &  $\log\fbrac{1/\confidence}\max\sbrac{\frac{1}{\approxerror},\frac{1}{\lipschitzconstant_{\wassmeas}^2},\lipschitzconstant_{\wassmeas}^2}$   &   $\frac{1}{\approxerror}\fbrac{\log\fbrac{\frac{\lipschitzconstant_\wassmeas}{\approxerror}}}^\weibull$\\
        \hline
        \end{tabular}
        \label{tab:slWassbaseBounds}
    \end{table}
\end{restatable}
\textbf{PAC Gurantee of \slwassbase.} We know (Equation~\eqref{equation:error_decompose}) that controlling the distance between the empirical measure and the sublinear summary is enough to control the distance estimation error if the empirical distribution concentrates to the true one. 
Lemma~\ref{Lemma; Concentration of Bounded-Tail Empirical Measure in Wasserstein Distance} of \citet{Bhat/NeurIPS/2019/ConcentrationRiskMeasuresThrough1DWasserstein} proposes the concentration guarantees of empirical measure in Wasserstein distance. Combining this result and Theorem~\ref{Theorem: slWassBase Works} yields estimation guarantees of $\estwassmeas$ w.r.t. true measure $\wassmeas$.

\begin{restatable}[PAC Guarantee of \slwassbase]{theorem}{slwassbasePAC}\label{Theorem: SWA Estimation Guarantee w.r.t. the True Measure}
Given $\size{\stream} = \wassmeassamplecount \geq \nthres{\lipschitzconstant_\wassmeas,\approxerror,\confidence} $ generated from a bounded-tail distribution $\wassmeas$, with bucket width $\bucketsize=\approxerror/2$, then \slCDFbase{} uses $\kthres{\lipschitzconstant_\wassmeas,\approxerror}$ space and outputs $\estwassmeas$ such that 
    $\mathbb{P}(\Wasserstein{1}{\wassmeas}{\estwassmeas} \leq \approxerror) \geq 1-\confidence\,.$
    This further shows that for a stream of size $n=\bigo{\epsilon^{-2}\log(1/\delta)}$ from two distributions $\wassmeas$ and $\wassmeastwo$, \slwassbase{} yields\vspace*{-.5em}
    \begin{align}
        |\Wasserstein{1}{\estwassmeas}{\estwassmeastwo} - \Wasserstein{1}{\wassmeas}{\wassmeastwo} |\leq 4\approxerror\,
    \end{align}
    with probability $1 - 2\confidence$. Here, $\epsilon \in (0,1/4]$ and $\delta \in (0,1/2]$. Here, $\subgaussianparameter \triangleq \max\sbrac{\subgaussianparameter_\wassmeas,\subgaussianparameter_\wassmeastwo}$, $\lipschitzconstant \triangleq \max\sbrac{\lipschitzconstant_\wassmeas,\lipschitzconstant_\wassmeastwo}$. We omit the constants in $\nthres{\lipschitzconstant_\wassmeas,\approxerror,\confidence}$ and $\kthres{\lipschitzconstant_\wassmeas,\approxerror}$ for simplicity.
        \begin{table}[H]
        \centering
        \begin{tabular}{c|cc}
        Tail Condition    &  $\nthres{\lipschitzconstant_\wassmeas,\approxerror,\confidence}$ &  $\kthres{\lipschitzconstant_\wassmeas,\approxerror}$\\
        \hline
        $\subgaussianparameter$-sub-Gaussian   &  ${\log\fbrac{1/\confidence}\max\sbrac{\frac{1}{\approxerror^2},\frac{1}{\lipschitzconstant^2},\lipschitzconstant^2}}$ &  ${\frac{\subgaussianparameter}{\approxerror}\log\fbrac{\frac{\lipschitzconstant}{\approxerror}}}$ \\
        $\weibull$-sub-Weibull    &  ${\log\fbrac{1/\confidence}\max\sbrac{\frac{1}{\approxerror^2},\fbrac{\log\frac{1}{\confidence}}^{2\weibull-1},\frac{1}{\lipschitzconstant^2},\lipschitzconstant^2}}$   
        &   ${\frac{1}{\approxerror}\fbrac{\log\fbrac{\frac{\lipschitzconstant}{\approxerror}}}^\weibull}$\\
        \hline
        \end{tabular}
        \label{tab:slWassbasePacBounds}
    \end{table}
\end{restatable}

\begin{remark}[Space, Time, and Communication Complexity of \slwassbase.]
    \slwassbase{} requires $\bigot{\frac{1}{\approxerror}}$ time, space, and communication cost to learn a distribution in Wasserstein distance. 
    Observe that the lower bound on $\wassmeassamplecount$ in Theorem~\ref{Theorem: SWA Estimation Guarantee w.r.t. the True Measure} ensures convergence $\empwassmeas$ to $\wassmeas$ in Wasserstein distance. This is the only term that has $\delta$-dependence as the rest of the algorithm is deterministic. \slwassbase's time, space and communication complexities are sublinear ($\widetilde{\Theta}(\sqrt{\wassmeassamplecount})$) compared to the number of samples required to ensure convergence of the empirical measure while retaining the same error.
\end{remark}





\subsection{Estimating TV Distance}\label{sec:estimateTV}

In this section, we introduce \slTVbase{} that yields an estimate of TV distance between two bounded-tail distributions with Lipschitz PDFs. \slTVbase{} leverages \slPDFbase{} to sublinearly learn the PDFs of the distributions, and then, estimates the TV distance as $L_1$ norm between them (Definition~\ref{Definition: TV Distance}).


\begin{algorithm}[ht!]
    \caption{{Sublinear TV Distance Estimator}:  $\slTVbase(\stream_{\wassmeas}, \stream_{\wassmeastwo} \overset{\iid}{\gets} \wassmeas, \wassmeastwo, \approxerror, \bucketsize)$}\label{Algorithm: Base TV Distance Approximation}
    \begin{algorithmic}[1]     \REQUIRE $\subgaussianparameter_\wassmeas, \subgaussianparameter_\wassmeastwo  \gets$ Subgaussian parameter of $\wassmeas$ and $\wassmeastwo$, $\bucketsize \gets$ bucket width
        \FOR{$\distribution \in \{\wassmeas, \wassmeastwo\}$}
        \STATE $\countersize_{\distribution} \gets \kthres{\approxerror,\lipschitzconstant,\confidence}$
        \STATE $\bucketsize_\distribution \gets \bthres{\approxerror,\lipschitzconstant}$
        \STATE $\{\pdf_{\hat{\distribution}_n}(i)\}_{i=1}^k = \{\hat{\pdf}_n(i)\}_{i=1}^k \gets \slPDFbase{}\fbrac{\distribution,\countersize_{\distribution},\bucketsize_\distribution}$
        \ENDFOR
            \STATE \textbf{return} $\TV{\estwassmeas}{\estwassmeastwo} \gets 0.5\sum_{i}\abs{\pdf_{\estwassmeas(i)}-\pdf_{\estwassmeastwo(i)}}\mathbf{1}\fbrac{\max\sbrac{\pdf_{\estwassmeas(i)},\pdf_{\estwassmeastwo(i)}>0}}$ 
    \end{algorithmic}
\end{algorithm}

\textbf{PAC Guarantee of \slTVbase.} \textit{We establish the estimation guarantee of $\slTVbase$ in three steps}. \underline{First}, we prove that if we choose the bucket width properly to discretise the support of $\wassmeas$, we obtain a control over the TV distance between true distribution and its bucketed version. \underline{Second}, we leverage the concentration inequalities of~\citet{Berend/SPLetters/2013/SharpEstimateTVConcentrationTrueMeasure} to control the TV distance between the bucketed distribution and the empirical bucketed distribution from the streamed samples. \underline{Finally}, we establish the approximation guarantee of \slPDFbase{} to control the TV distance between the empirical bucketed distribution and its sublinear summary. 

\textbf{Step 1: Setting the Bucket Width.} 
{Standard rules~\citep{ScottHistogramBins1979,FreedmanDiaconis/ZWVG/1981/HistogramBinWidthForDensityEstimation} of choosing optimal bucket width of a histogram depends either on rule of thumbs or crucially depends on parameters such as sample size $\wassmeassamplecount$, $\int_{-\infty}^{\infty} \pdf'(x)^2 \, dx$ etc., which are difficult to estimate for an unknown distribution arriving in a stream. We propose a bucketing technique depending on the Lipschitz constant of the true PDF and the sub-Gaussian parameter of the true distribution to tune the bucket width a priori. {Proof is in Appendix~\ref{Appendix: Proof of STVA}}}

\begin{restatable}[Bucket Width for TV Distance Estimation]{theorem}{bucketingtv}\label{Theorem: bucket width for Subgaussian TV}
                Given a bounded-tail distribution $\wassmeas$ with $\lipschitzconstant_\wassmeas$-Lipschitz PDF and a corresponding bucketed measure $\bucketwassmeas{\bucketsize}$, if we fix the bucket width $\bucketsize = \bthres{\approxerror,\lipschitzconstant}$, we have $\TV{\wassmeas}{\bucketwassmeas{\bucketsize}} \leq \epsilon$.
        \begin{table}[H]
            \centering
            \begin{tabular}{c|cc}
                Tail Condition    & $\subgaussianparameter_\wassmeas$-sub-Gaussian  & $\weibull$-sub-Weibull  \\
                \hline
                $\bthres{\approxerror,\lipschitzconstant_\wassmeas}$    & $\frac{\approxerror}{\subgaussianparameter_\wassmeas\lipschitzconstant_\wassmeas\sqrt{\log\fbrac{2/\approxerror}}}$  &  $\frac{\approxerror}{\lipschitzconstant_\wassmeas\fbrac{\log\fbrac{2\constant_\weibull/\approxerror}}^{\weibull}}$ \\
                \hline
            \end{tabular}
            \label{tab:TVBucketingBounds}
        \end{table}
\end{restatable}

\textbf{Step 2: Concentration of Bucketed Empirical Distribution in TV.} For continuous measures, assumption on the structure of the distribution is necessary to establish meaningful convergence rates in TV distance~\citep{Diakonikolas/HandbookOfBigData/2016/LearningStructuredDistributions}. We establish a convergence result for bucketed measure $\bucketwassmeas{\bucketsize}$ generated from a bounded-tail measure $\wassmeas$ that possibly has infinite buckets in its support (Appendix~\ref{Appendix: Proof of STVA}).

\begin{restatable}[Concentration in TV over Infinite Buckets]{lemma}{TVdistanceconc}\label{lemma: Conc of TV distance}
Let $\empwassmeas$ be an empirical measure generated from a discrete bucketed measure $\wassmeas_{\bucketsize}$ corresponding to a bounded-tail distribution $\wassmeas$, and, $\approxerror,\confidence \in \fbrac{0,1}$. Then, for $\wassmeassamplecount \geq \nthres{\approxerror,\bucketsize,\confidence}$, $\mathbb{P}\tbrac{\TV{\wassmeas}{\empwassmeas} \geq \approxerror} \leq \confidence$. Here, $\Gamma(\cdot)$ denotes the Gamma function~\citep{GammaFunction}.
    \begin{table}[H]
            \centering
                \renewcommand{\arraystretch}{1.4}
            \begin{tabular}{c|cc}
                Tail Condition    & $\subgaussianparameter_\wassmeas$-sub-Gaussian  & $\weibull$-sub-Weibull  \\
                \hline
                $\nthres{\approxerror,\bucketsize,\confidence}$    & $\constant\approxerror^{-2}\max\sbrac{\frac{4\subgaussianparameter_\wassmeas\sqrt{\pi}}{\bucketsize},\log\fbrac{1/\confidence}}$  &  $\constant\approxerror^{-2}\max\sbrac{\frac{2\constant_\weibull}{\bucketsize}\Gamma(1+\alpha),\log\fbrac{1/\confidence}}$ \\
                \hline
            \end{tabular}
            \label{tab:TvConcBounds}
        \end{table}
\end{restatable}

\textbf{Step 3: Approximation Guarantee of \slPDFbase.} Theorem~\ref{Theorem: slPDFbase works} shows that \slPDFbase{} learns a pointwise approximation to the PDF of any bucketed empirical measure. However, to extend this guarantee to the TV distance, we need to show that the the sum of errors over the entire support is small. Note that bounded-tail distributions have small mass in the tails (i.e. in most of the buckets). Hence, it suffices to control the error in learning PDF of the `heavier' buckets to control the error in estimating the TV distance. The proof is in Appendix~\ref{Appendix: Proof of STVA}.

\begin{restatable}[Learning Error of \slPDFbase{}]{theorem}{slTVbaseWorks}\label{Theorem: Tv Distance Approx Algo Works}
    If \slPDFbase{} accesses a stream of length $\size{\stream} = \wassmeassamplecount \geq \nthres{\approxerror,\confidence}$ from a bounded-tail distribution, and uses $\kthres{\approxerror,\confidence}$ buckets to output a sublinear summary $\pdf_{\estwassmeas}$, then with probability $1-\delta$, $\TV{\estwassmeas}{\empbucketwassmeas{\bucketsize}} \leq \norm{\pdf_{\estwassmeas} - \pdf_{\empbucketwassmeas{\bucketsize}}}_1 \leq \approxerror$. 
\vspace{-1.5em}
    \begin{table}[H]
        \centering
            \renewcommand{\arraystretch}{1.8}
        \begin{tabular}{c|cc}
        Tail Condition    &  $\nthres{\approxerror,\confidence}$ &  $\kthres{\approxerror,\bucketsize}$\\
        \hline
        $\subgaussianparameter_\wassmeas$-sub-Gaussian   &  \blue{$\frac{\constant}{\approxerror}\log\fbrac{\frac{1}{\confidence}}$} &  $\basemglengthpdf$ \\
        \hline
        $\weibull$-sub-Weibull    &  $\frac{\constant}{\approxerror}\log\fbrac{\frac{1}{\confidence}}$   &   $\ceil*{\frac{\constant}{\bucketsize}\fbrac{\log\fbrac{\frac{6}{\approxerror}}}^\weibull}$\\
        \hline
        \end{tabular}
        \label{tab:slTVbaseBounds}
    \end{table}
\end{restatable}

These results together yield the PAC guarantee for \slTVbase.

\begin{restatable}[PAC Guarantee of \slTVbase]{theorem}{slTVbasePAC}\label{Theorem: PAC of SLTV}
Given $\epsilon \in (0,1/6]$, $\delta \in (0,1/4]$, a stream of size $\wassmeassamplecount\geq\nthres{\approxerror,\lipschitzconstant,\confidence}$ from two bounded-tail and $\ell$-Lipschitz distributions $\wassmeas$ and $\wassmeastwo$, and bucket width $\bucketsize = \bthres{\approxerror,\lipschitzconstant,\confidence}$, \slTVbase{} uses $\kthres{\approxerror,\lipschitzconstant,\confidence}$ space and
    \begin{align}
       \mathbb{P}\fbrac{ |\TV{\estwassmeas}{\estwassmeastwo} - \TV{\wassmeas}{\wassmeastwo} |\geq 6\approxerror}\leq 4\delta\,.
    \end{align}
    Here, $\subgaussianparameter \triangleq \max\sbrac{\subgaussianparameter_\wassmeas,\subgaussianparameter_\wassmeastwo}$, $\lipschitzconstant \triangleq \max\sbrac{\lipschitzconstant_\wassmeas,\lipschitzconstant_\wassmeastwo}$. We omit the constants in $\nthres{\lipschitzconstant_\wassmeas,\approxerror,\confidence}$ and $\kthres{\lipschitzconstant_\wassmeas,\approxerror}$ for simplicity.
    \begin{table}[H]
        \centering
        \renewcommand{\arraystretch}{1.5}
        \begin{tabular}{c|ccc}
        Tail Condition    &  $\nthres{\approxerror,\lipschitzconstant,\confidence}$ &  $\kthres{\approxerror,\lipschitzconstant}$ & $\bthres{\approxerror,\lipschitzconstant}$\\
        \hline
        $\subgaussianparameter$-sub-Gaussian   &  ${\approxerror^{-2}\max\sbrac{\frac{\subgaussianparameter^2\lipschitzconstant
    \log\fbrac{1/\approxerror}}{\approxerror},\log\fbrac{\frac{1}{\confidence}}}}$ &  $\frac{\subgaussianparameter^2\lipschitzconstant}{\approxerror}\log\fbrac{\frac{1}{\approxerror}}$ & $\frac{\approxerror}{\subgaussianparameter\lipschitzconstant\sqrt{\log\fbrac{2/\approxerror}}}$  \\
        \hline
        $\weibull$-sub-Weibull    &  ${\approxerror^{-2}\max\sbrac{\frac{\lipschitzconstant\fbrac{\log\fbrac{1/\approxerror}}^\weibull}{\approxerror}\Gamma(1+\alpha),\log\fbrac{\frac{1}{\confidence}}}}$   &   $\frac{\lipschitzconstant}{\approxerror}\fbrac{\log\fbrac{\frac{1}{\approxerror}}}^{2\weibull}$ &  $\frac{\approxerror}{\lipschitzconstant\fbrac{\log\fbrac{2\constant_\weibull/\approxerror}}^{\weibull}}$\\
        \hline
        \end{tabular}
    \end{table}
\end{restatable}

\begin{remark}[Space, Time, Communication Complexity of \slTVbase]
    \slTVbase{} requires $\bigot{\frac{1}{\approxerror}}$ time, space, and communication complexity per round to estimate the TV distance between two bounded-tail distributions\footnote{Note that for sub-Gaussians $\alpha = \frac{1}{2}$.}. Observe that the lower bound on $\wassmeassamplecount$ ensures convergence of $\estwassmeas$ to $\wassmeas$ in TV distance (Theorem~\ref{Theorem: PAC of SLTV}). This is the only term with $\delta$-dependence as the rest of the algorithm is deterministic. Thus, \slTVbase{} achieves $\widetilde{\Theta}(\wassmeassamplecount^{1/3})$ time, space, and communication complexity with respect to \#samples required for convergence of empirical measure while retaining the order of error.
\end{remark}

Note that the bounds in Theorem~\ref{Theorem: PAC of SLTV} ensures that $\approxerror$ is of the order $\wassmeassamplecount^{-1/3}$, which ensures that the bucket width is of the order $\wassmeassamplecount^{-1/3}$ and the error in TV distance is of the order $\wassmeassamplecount^{-1/3}$. Our result is consistent with those of other standard histogram bucket width rules, which chooses bucket width of order $\wassmeassamplecount^{-1/3}$ and ensures the integrated mean squared error to be of order $\wassmeassamplecount^{2/3}$~\citep{ScottHistogramBins1979, FreedmanDiaconis/ZWVG/1981/HistogramBinWidthForDensityEstimation}.

\begin{remark}[Further Implications of Theorem~\ref{Theorem: Tv Distance Approx Algo Works}]\label{Remark: L1 frequency estimation}
We observe that Theorem~\ref{Theorem: Tv Distance Approx Algo Works} does not only yield a guarantee on TV distance estimation error but also have two further implications, which might be of general interest.

1. \textit{$\ell_1$-estimation of True Frequencies:} A closer look into the result shows that \slPDFbase{} can also yield an  $\ell_1$ approximation of the true frequencies ($\{\truefrequency{i}\}_i$) corresponding to any probability distribution when the tails are sufficiently bounded. 

2. \textit{Estimation of any Integral Probability Metric (IPM):}  As our algorithms yield pointwise estimates of both the PDF and CDF of a distribution, these methods can be used to estimate any Integral Probability Metric (IPM), such as Dudley metric, Maximum Mean Discrepancy (MMD), Kolmogorov distance~\citep{Sriperumbudur/EJS/2012/EmpiricalEstimationIPMContinuousTV}.
\end{remark}


\input{Sections_PODS/5.1_Lower_Bounds}

\input{icml2024/Sections/5_Experiments}

%% file: icml2024/Figures/summary_figures.tex

\begin{figure*}[t!]
\centering
\includegraphics[width=0.24\textwidth]{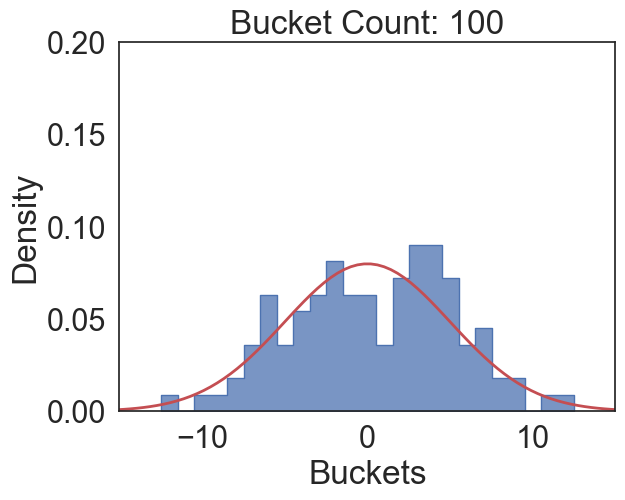}\hfill
\includegraphics[width=0.24\textwidth]{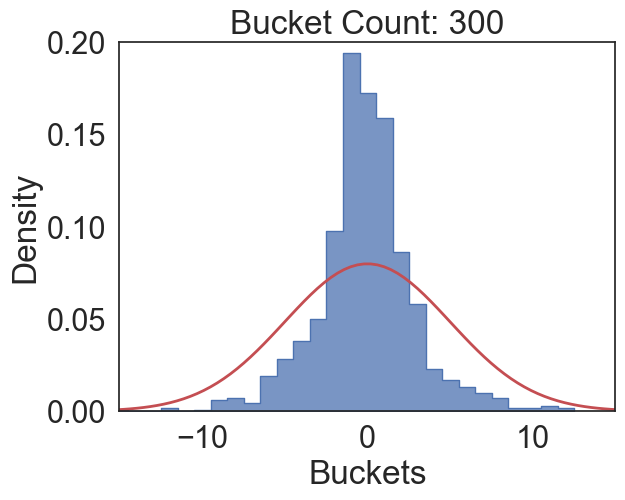}\hfill
\includegraphics[width=0.24\textwidth]{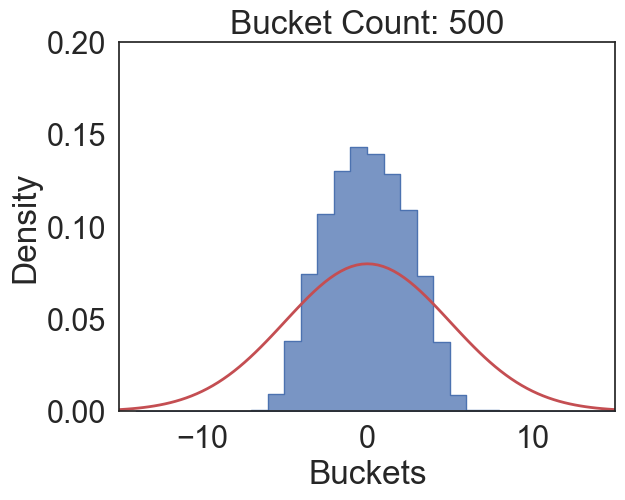}\hfill
\includegraphics[width=0.24\textwidth]{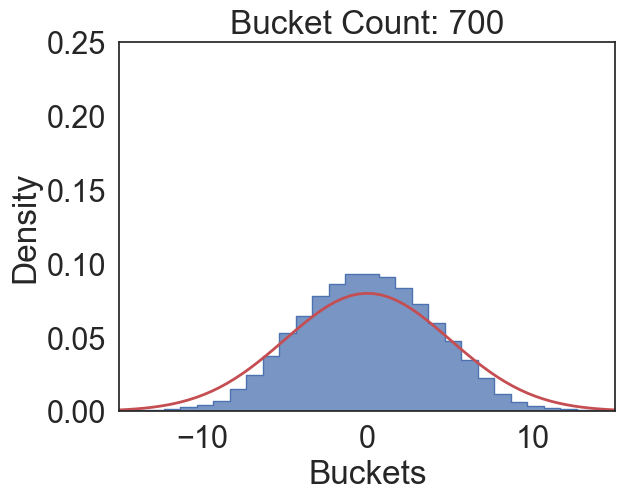}\vspace{-1em}
\caption{Evolution of summaries constructed using $10^5$ samples from $\normal(0,5)$. We set $\bucketsize = 0.05$ and bucket numbers $100,300,500,700$.}\label{Fig: pdf learn}
\end{figure*}

%% file: Sections_PODS/5.1_Lower_Bounds.tex
\subsection{Tightness of Space Complexity}
We compare upper bounds on space complexities of \slPDFbase{} and \slCDFbase{} to the existing lower bounds for finite universe of streams. Frequency estimation have been extensively studied in this setting. 
Theorem~\ref{thm:lower_bound} provides a $\bigomega{\approxerror^{-1}}$ lower bound on space complexity of any non-trivial solution, i.e. not storing the entire stream, or counts of all elements in the universe, to the frequency estimation problem with respect to the $\ell_\infty$-norm.

\begin{theorem}[Lower Bound of Frequency Estimation in $\ell_\infty$-norm~\citep{Chakrabarti_2024}]\label{thm:lower_bound}
    Given a stream $\stream$ of length $\streamlength$ coming from a universe $\streamuniverse$ of size $\streamuniversesize$, any algorithm that produces a $\fbrac{\approxerror\streamlength}$-$\ell_\infty$ approximation of the frequency vector $\sbrac{\truefrequency{i}}_{i \in \tbrac{\streamuniversesize}}$ as $\sbrac{\estimatedfrequency{i}}_{i \in \tbrac{\streamuniversesize}}$ must use $\bigomega{\min\fbrac{\streamlength,\streamuniversesize,\approxerror^{-1}}}$ space.
\end{theorem}

Our framework provides a $\bigot{\approxerror^{-1}}$ approximation of the true frequencies in $\ell_1$-norm for any bounded-tail distribution with both infinite and finite supports, i.e. universe of streams (ref. Remark~\ref{Remark: L1 frequency estimation})\footnote{Note that while our techniques consider the data arrives from a distribution rather than arbitrarily from a universe (as in Theorem~\ref{thm:lower_bound}), it suffices if the underlying data are generated from a distribution and the order of their arrivals is arbitrary.}. Hence, Theorem~\ref{thm:lower_bound} indicates that \textit{our algorithms are tight in terms of space complexity}.




%% file: icml2024/Sections/5_Experiments.tex
\section{Experimental Analysis}
\label{section: experiments_main}
Our aim is to understand\textit{ whether \slwassbase{} and \slTVbase{} yield PAC estimates of Wasserstein and TV distances between sub-Gaussian distributions with sublinear space, i.e. number of buckets, with respect to the number of samples.}

\textbf{Setup.} We compute distances between two Gaussian distributions $\mathcal{N}(0,5)$ and $\mathcal{N}(1,5)$, where $10^5$ samples from each of them arrive through $\clients=10$ sources. True TV and Wasserstein distances between them are $0.0797$ and $1.00$, respectively. We set the bucket width to $\bucketsize=0.05$ and increase the number of buckets as $\{100, 150,\ldots,1000\}$. We run each of the experiments $50$ times. Finally, for \slwassbase{}, we report the estimated Wasserstein distance and the learning error of \slCDFbase, i.e. $\Wasserstein{p}{\empwassmeas}{\estwassmeas}$, in Figure~\ref{Fig: Wasserstein Synthetic}. Similarly, for \slTVbase{}, we report the estimated TV distance and the learning error of \slPDFbase, i.e. $\TV{\empwassmeas}{\estwassmeas}$, in Figure~\ref{Fig: TV Synthetic}. We report the true distances in each case as the dotted horizontal lines. 

\textbf{Results.} 

1. \textbf{Accuracy and Sublinearity of \slwassbase.} The estimation error of \slwassbase{} and learning error becomes negligible as the bucket number reaches to $750$ while using $10^5$ samples. This is better than the theoretical upper limit of buckets, i.e. $1040$, suggested by our results. 

2. \textbf{Accuracy and Sublinearity of \slTVbase.} The estimation error of \slTVbase{} and learning error becomes negligible as \#buckets also reaches to $750$. Thus, \textit{the results demonstrate that both \slTVbase{} and \slwassbase{} yield accurate estimates of TV and Wasserstein distances while using only 
sublinear space w.r.t. \#samples.}

%% file: Sections_PODS/4_ApplicationAuditing.tex
\subsection{Applications: Auditing Fairness and Privacy}\label{sec:audits}
Auditing fairness~\citep{DBLP:conf/aaai/2021/GhoshBM21/Justicia,DBLP:conf/aaai/2022/GhoshBM/FVGM,activeaudit} and privacy~\citep{nasr2023tight,steinke2024privacy,koskela2024auditingdifferentialprivacyguarantees,annamalai2024nearly,azize2025some} of Machine Learning (ML) models is an important and increasingly studied question for developing trustworthy ML. 
Here, we deploy \slwassbase{} and \slTVbase{} to estimate the fairness and privacy of ML models trained on real-world data, respectively. The details are provided in~\Cref{app:Experiments}.


\input{icml2024/Figures/performance_plots}

%% file: icml2024/Figures/performance_plots.tex
\begin{figure*}[t!]
\centering
\begin{minipage}{0.45\textwidth}
\includegraphics[width=\linewidth]{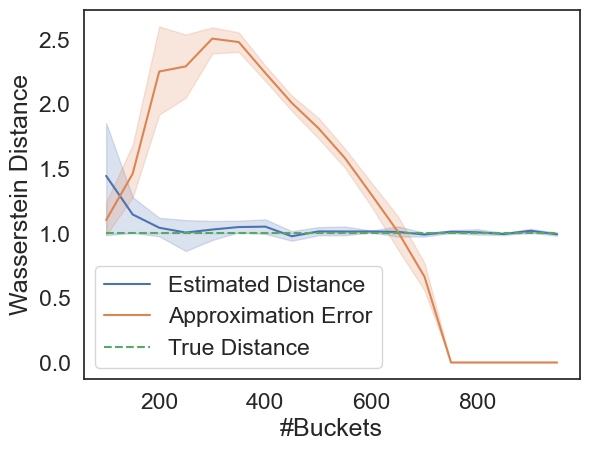}\vspace*{-1em}
\caption{Performance of \slwassbase{} with $\normal(0,5)$, $\normal(1,5)$ and $\bucketsize = 0.05$}\label{Fig: Wasserstein Synthetic}
\end{minipage}\hfill
\begin{minipage}{0.45\textwidth}
\includegraphics[width=\linewidth]{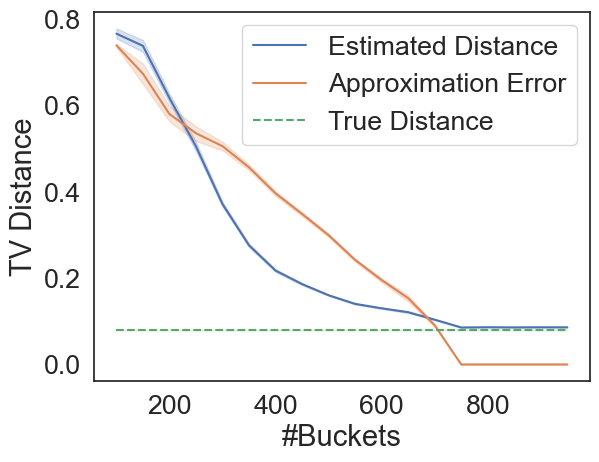}\vspace*{-1em}
\caption{Performance of \slTVbase{} with $\normal(0,5)$, $\normal(1,5)$ and $\bucketsize = 0.05$}\label{Fig: TV Synthetic}
\end{minipage}
\end{figure*}
\begin{figure*}
\begin{minipage}{0.33\textwidth}
\includegraphics[width=\linewidth]{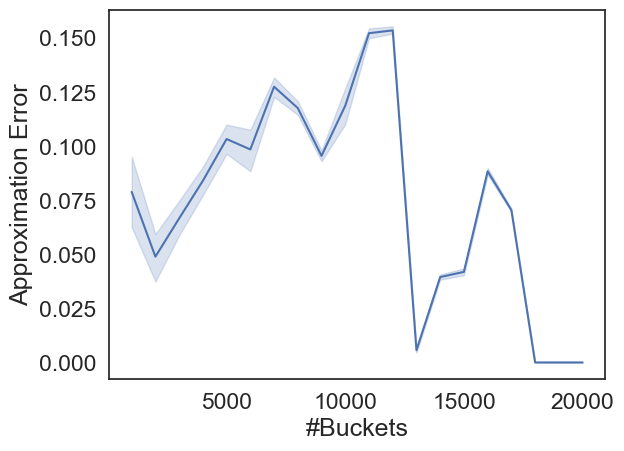}\vspace*{-1em}
\caption{Auditing with \slwassbase{} on regression output of \acsincome.}\label{Fig: ACS Income Regression}
\end{minipage}\hfill
\begin{minipage}{0.33\textwidth}
\includegraphics[width=\linewidth]{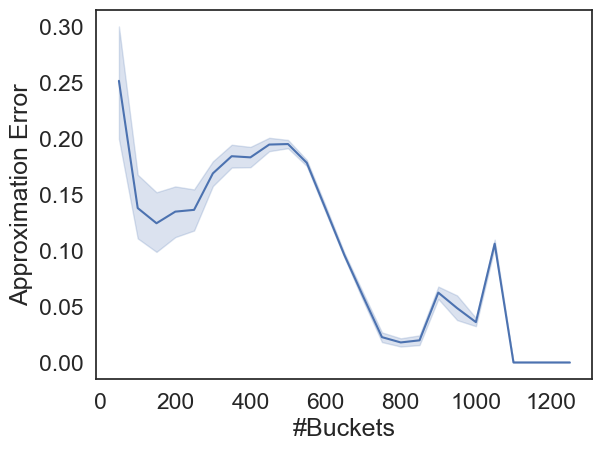}\vspace*{-1em}
\caption{Auditing with \slwassbase{} on classification output of \acsincome.}\label{Fig: ACS Income Classification}
\end{minipage}\hfill
\begin{minipage}{0.33\textwidth}
\includegraphics[width=\linewidth]{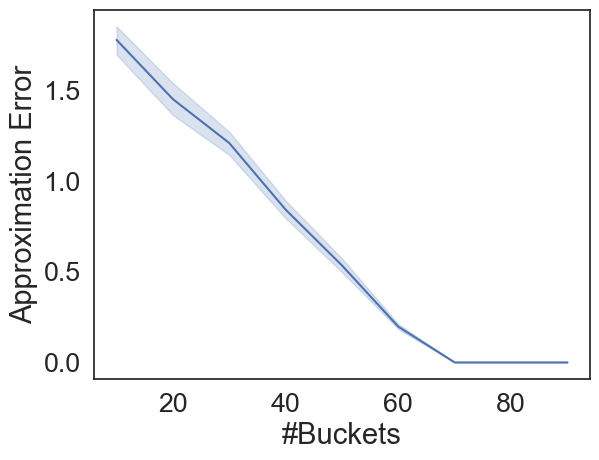}\vspace*{-1em}
\caption{Privacy Auditing of logistic regression on MNIST.}\label{Fig: Noisy Regression}
\end{minipage}\vspace*{-1em}%
\end{figure*}

%% file: Sections_PODS/6_Conclusion.tex
\section{Discussions and Future Works}
We propose an algorithmic framework to learn \textit{mergeable} and \textit{sublinear} summaries of discrete and continuous, sub-Gaussian and sub-Weibull distributions while data streams arrive from a single or multiple sources. We show that the computed mergeable sublinear summarisers are universal estimators with respect to Wasserstein distance. We establish the first sublinear time, space, and communication algorithms to estimate TV and Wasserstein distances over continuous and discrete (with infinite support) distributions. We also note that our framework can be used to estimate any $\ell_p$-norm based distance between distributions with sublinear resources. However, our work is constrained to distributions over scalars. A future direction is to study these problems for multi-variate distributions. 

%% file: Sections_PODS/Appendix.tex
\input{Sections_PODS/Appendix/1_Megrable_Misra_Gries}

\input{Sections_PODS/Appendix/2_Empirical_Measure}

\input{Sections_PODS/Appendix/3_Distribution_Summaries}

\input{Sections_PODS/Appendix/4_Wasserstein_Approximation}

\input{Sections_PODS/Appendix/4.5_TV_Distance_Estimation}
\input{Sections_PODS/Appendix/5_Bucketing_TV}

\input{Sections_PODS/Appendix/6_applications}
\input{Sections_PODS/Appendix/9_Empirical_Tail_Bounds}

\input{Sections_PODS/Appendix/7_useful_results}
\input{Sections_PODS/Appendix/X_Experiments}
\input{Sections_PODS/Appendix/8_Large_Figures}

%% file: Sections_PODS/Appendix/1_Megrable_Misra_Gries.tex
\newpage
\section{Mergeable Misra Gries Algorithm}\label{Appendix: MMG Algo Exposition}

\begin{algorithm}[ht!]
\caption{\mergablemisragries~\citep{Efficient_Mergable_Misra_Gries}}\label{Algo: Mergable Misra-Gries}
\begin{algorithmic}[1]{
\STATE \textbf{\initializestream(}{k}\textbf{)}
    \STATE $T \gets \emptyset$ \qquad\qquad\qquad\qquad\qquad $\triangleright$~~$T$ is the set of items for which counters are assigned
    \STATE $k \gets k$
    \STATE $k^* \gets \frac{k}{2}$
\STATE

\STATE \textbf{\updatestream(}{$i, \Delta$}\textbf{)}
    \IF{$\streamelement{i} \in T$}
        \STATE $c(\streamelement{i}) \gets c(\streamelement{i}) + \Delta$
    \ELSIF{$|T| < k$}
        \STATE $T = T \cup \{\streamelement{i}\}$
        \STATE $c(\streamelement{i}) \gets c(\streamelement{i}) + \Delta$
    \ELSE
        \STATE DecrementCounters()
    \ENDIF
    \IF{$\Delta \geq c_{k^*}$} 
        \STATE  \label{linediep1}$T = T \cup \{\streamelement{i}\}$ \qquad\qquad\qquad $\triangleright$~~$|T| \leq k^*+1 \leq k$ after this line
        \STATE  \label{linedie2} $c(\streamelement{i}) \gets c(\streamelement{i}) + \Delta - c_{k^*}$
    \ENDIF

\STATE
\STATE \textbf{DecrementCounters()}{} \label{diediedie}
    \STATE \label{notation} // Notation: Let $c_{k^*}$ be the $k^*$-th largest value, counting multiplicity,
    \STATE// in the multiset $\{c(\streamelement{j}) \colon \streamelement{j} \in T\}$.
    \FOR{$\streamelement{j} \in T$:}
        \STATE $c(\streamelement{j}) = c(\streamelement{j}) - c_{k^*}$ \label{theline}
        \STATE \textbf{if } $c(\streamelement{j}) \leq 0$:
        \STATE $T = T \setminus \{\streamelement{j}\}$ \label{die2} \label{line:unassign}
    \ENDFOR
\STATE

\STATE \textbf{\estimatestream(}{$i$}\textbf{)}
\IF{$\streamelement{i} \in T$}
\RETURN  $c(\streamelement{i})$
\ELSE
\RETURN  0
\ENDIF
\STATE

\STATE \textbf{\estimatecumulant(}{$\streamelement{i}$}\textbf{)}
    \STATE $c\fbrac{\streamelement{i}} \leftarrow \estimatestream\fbrac{\streamelement{i}}$
    \RETURN  $\estimatedcumulant{i} \leftarrow \sum_{\streamelement{j} \leq \streamelement{i}} C_j$
\STATE

\STATE \textbf{\merge(}{$T_2$}\textbf{)}
    \FOR{$\streamelement{i} \in T_2$}
        \STATE  Update$_{T_1}(\streamelement{i}, c(\streamelement{i}))$
        \RETURN $T_1$
    \ENDFOR
}
\end{algorithmic}
\end{algorithm}

First, we outline the proof for the Lemma~\ref{Theorem: Mergeable Misra Gries}, restated here for ease of reading.

\MMGGuarantee*

The proof of the result follows that of Theorem~2 in~\cite{Efficient_Mergable_Misra_Gries}. The change is in the Lemma~4. We provide the proof for the updated statement here. The notation is kept same as that of the original work for ease of reading.

As in the original paper, we define $\frequencysum{l} = \sum_{i = 1}^l \streamweight{j}$, $\countersum{l} = \sum_{i = 1}^\countersize \counter{i}$, and $\errorsum{l} = \sum_{i \in [\streamuniversesize]} \truefrequency{i,l} - \estimatedfrequency{i,l}$

\begin{lemma}\label{Lemma: Fixed Lemma 4}
    $\errorsum{n} \leq \fbrac{\frequencysum{n} - \countersum{n}}/\fbrac{\countersize - \counterthreshold}$
\end{lemma}

\begin{proof}
    As in the original proof, we proceed by proof by induction, the case for $l = 0$ is true as $\frequencysum{o} = \countersum{0} = \errorsum{0} = 0$.

    Suppose the hypothesis holds for some $l-1$, i.e.:
    \begin{align*}
        \errorsum{l-1} \leq \fbrac{\frequencysum{l-1} - \countersum{l-1}}/\fbrac{\countersize - \counterthreshold}
    \end{align*}
    If the $l$-th element in the stream does not cause \decrement{} to be called, we have $\errorsum{l} = \errorsum{l-1}$, $\countersum{l} = \countersum{l-1} + \streamweight{l}$, $\frequencysum{l} = \frequencysum{l-1} + \streamweight{l}$. Hence ,we have:
    \begin{align*}
        \errorsum{l} = \errorsum{l-1} \leq \fbrac{\frequencysum{l-1} - \countersum{l-1}}/\fbrac{\countersize - \counterthreshold} = \fbrac{\frequencysum{l} - \countersum{l}}/\fbrac{\countersize - \counterthreshold}
    \end{align*}

    The case when \decrement{} is called is more interesting, we have 
    \begin{align}
    \errorsum{l} = \errorsum{l-1}+\deccounter\label{Eq: Error Sum Bound}
    \end{align}
    Here, we know that at $l-1$-th step, $\fbrac{\countersize - \counterthreshold}$ counters had value $\geq \deccounter$. Therefore, they retain some non-negative value after being reduced by $\deccounter$, while the remaining $\counterthreshold$ counters have value $0$. Hence, we have $\countersum{l} \leq \countersum{l-1} - \fbrac{\countersize - \counterthreshold}\deccounter$.
    \begin{align}
        \nonumber&\frequencysum{l} - \countersum{l}\\\nonumber
        =&\frequencysum{l-1} + \streamweight{l} - \countersum{l}\\\nonumber
        \geq&\frequencysum{l-1} + \streamweight{l} -\countersum{l-1} + \fbrac{\countersize - \counterthreshold}\deccounter\\
        \geq&\frequencysum{l-1} -\countersum{l-1} + \fbrac{\countersize - \counterthreshold}\deccounter\label{Eq: Counter deviation bound}
    \end{align}
    Now, combining Equations~\eqref{Eq: Error Sum Bound} and~\eqref{Eq: Counter deviation bound}, we have:
    \begin{align*}
        \errorsum{l} = \errorsum{l-1} + \deccounter
        \leq \frac{\frequencysum{l-1} - \countersum{l-1}}{\fbrac{\countersize - \counterthreshold}} + \deccounter
        = \frac{\frequencysum{l-1} - \countersum{l-1} + \fbrac{\countersize - \counterthreshold}\deccounter}{\fbrac{\countersize - \counterthreshold}}
        \leq \frac{\frequencysum{l} - \countersum{l}}{\fbrac{\countersize - \counterthreshold}}
    \end{align*}
\end{proof}

The second part of the proof of Theorem~2 in the original paper follows similarly under the updated statement of Lemma~\ref{Lemma: Fixed Lemma 4}. Giving us the final result of Lemma~\ref{Theorem: Mergeable Misra Gries}.

To establish the estimation guarantee of \estimatecumulant{}, we need a simple corollary of the Lemma~\ref{Theorem: Mergeable Misra Gries}:

\begin{corollary}[\mergablemisragries{} Estimation Guarantees]\label{Corollary: Mergeable Misra Gries k/4}
    For $j \in \{1,2\}$, given streams $\stream_j$ being stored in two separate sets of counters of size $\countersize$, the merged summary satisfies:
    \begin{align*}
        0 \leq \truefrequency{i} - \estimatedfrequency{i} \leq \frac{4~\residualfrequency{\frac{\countersize}{4}}}{\countersize}
    \end{align*}
    for all $j \leq \counterthreshold$, where $\residualfrequency{j}$ denotes the sum frequency of all but $j$ of the most frequent items.
\end{corollary}

\begin{proof}
    We fix $\counterthreshold = \frac{\countersize}{2}$, and $j = \frac{\countersize}{4}$. These values satisfy the criteria of Theorem~\ref{Theorem: Mergeable Misra Gries}, i.e. $\frac{\countersize}{2} = \Omega(\countersize)$ and $j < \counterthreshold$. Plugging in these values gives the result.
\end{proof}

We now prove the Lemma~\ref{Lemma: Estimate Cumulant Works}:


\begin{proof}
    We denote by $\topk{\countersize}$ the set of $\countersize$ elements with highest true frequency $\truefrequency{i}$. Then, we have:
    \begin{align*}
        \truecumulant{i} - \estimatedcumulant{i}
        =& \sum_{j \leq i} \truefrequency{j} - \estimatedfrequency{j}\\
        =& \sum_{\substack{j \leq i \\ j \in \topk{\frac{\countersize}{4}}}} \truefrequency{j}  - \estimatedfrequency{j} + \sum_{\substack{j \leq i \\ j \notin \topk{\frac{\countersize}{4}}}} \truefrequency{j}\\
        \leq& \sum_{j \in \topk{\frac{\countersize}{4}}} \frac{4~\residualfrequency{\frac{\countersize}{4}}}{\countersize} + \sum_{j \notin \topk{\frac{\countersize}{4}}} \truefrequency{j}\\
        \leq& ~2~\residualfrequency{\frac{\countersize}{4}}
    \end{align*}
    Where the first inequality uses Corollary~\ref{Corollary: Mergeable Misra Gries k/4}, and the last inequality follows from the definition of $\residualfrequency{\countersize}$.
\end{proof}

%% file: Sections_PODS/Appendix/2_Empirical_Measure.tex




\section{Properties of Empirical Measure}\label{app:empiricaldist}

In this section we provide the proofs for Theorems~\ref{Theorem: Empirical Measure is Subgaussian},~\ref{theorem: Empirical Distribution is SubWeibull},~and~\ref{Theorem: Empirical Measure is Bi-Lipschitz}. We restate the theorems for easy of reading. We introduce the multiplicative Chernoff bound that will be relevant to our analysis.

\begin{lemma}[Multiplicative Chernoff Bound\citep{Mitzenmacher_Upfal_2005}]\label{Lemma: Multiplicative Chernoff Bound}
    Given i.i.d. random variables $X_1,X_2,...,X_n$ where $\Pr[X_i = 1] \leq p$ and, define $X = \frac{1}{n}\sum_{i \in [n]} X_i$. Then, we have:
    \begin{align}
    \Pr[X \geq (1+\approxerror) p] &\leq \exp{\fbrac{-\frac{n\approxerror^2p}{3}}} & 0 \leq \approxerror <1\\
    \end{align}
\end{lemma}

\EmpircalMeasureSubGaussian*

\begin{proof}
    To capture the tail behaviour w.r.t. some $\tail \leq \threshold$ for each sample, we define random variables $\sequence{Z}{\streamlength}$ as:
    \begin{align*}
        Z_i &= \begin{dcases}
           1 & X_i \geq \tail\\
           0 & X_i < \tail 
        \end{dcases}
    \end{align*}
    We denote by $Z = \frac{1}{\streamlength}\sum_{i \in \tbrac{\streamlength}} Z_i$. Then, we have for the empirical measure $\empwassmeas$,
    \begin{align*}
        \Pr_{X \sim \empwassmeas} \tbrac{X \geq \tail} = Z
    \end{align*}
    Furthermore, we have by~\Cref{Def: Subgaussianity}:
    \begin{align*}
        \Pr\tbrac{Z_i = 1} \leq 2\exp\fbrac{-\frac{\tail^2}{\subgaussianparameter^2}} 
    \end{align*}
    Then, by Lemma~\ref{Lemma: Multiplicative Chernoff Bound}, we have:
    \begin{align*}
        \Pr\tbrac{Z \geq 2\exp\fbrac{-\frac{\tail^2}{\subgaussianparameter^2}}} \leq \Pr\tbrac{Z \geq 3\exp\fbrac{-\frac{\tail^2}{\subgaussianparameter^2}}}\leq \exp\fbrac{-\frac{\streamlength\exp\fbrac{-\frac{\tail^2}{\subgaussianparameter^2}}}{6}} \leq \confidence
    \end{align*}
    By setting $\constant = 6$, we get the desired result.
\end{proof}

Now, we prove Theorem~\ref{theorem: Empirical Distribution is SubWeibull}, restated here for ease of reading:

\EmpircalMeasureSubWeibull*

\begin{proof}
    To capture the tail behaviour w.r.t. some $\tail \leq \threshold$ for each sample, we define random variables $\sequence{Z}{\streamlength}$ as:
    \begin{align*}
        Z_i &= \begin{dcases}
           1 & X_i \geq \tail\\
           0 & X_i < \tail 
        \end{dcases}
    \end{align*}
    We denote by $Z = \frac{1}{\streamlength}\sum_{i \in \tbrac{\streamlength}} Z_i$. Then, we have for the empirical measure $\empwassmeas$,
    \begin{align*}
        \Pr_{X \sim \empwassmeas} \tbrac{X \geq \tail} = Z
    \end{align*}
    Furthermore, we have:
    \begin{align*}
        \Pr\tbrac{Z_i = 1} \leq \constant_\weibull\exp\fbrac{-\tail^{1/\weibull}}
    \end{align*}
    Then, by Lemma~\ref{Lemma: Multiplicative Chernoff Bound}, we have:
    \begin{align*}
        \Pr\tbrac{Z \geq 1.5\constant_\weibull\exp\fbrac{\tail^{1/\weibull}}} \leq \exp\fbrac{-\frac{\streamlength\constant_\weibull\exp\fbrac{-\tail^{1/\weibull}}}{12}} \leq \exp\fbrac{-\frac{\streamlength\constant_\weibull\exp\fbrac{-\sqrt[\weibull]{\threshold}}}{12}} \leq \delta
    \end{align*}
\end{proof}

Next, we prove the Theorem~\ref{Theorem: Empirical Measure is Bi-Lipschitz}. For that purpose, we need the following lemma.

\begin{lemma}[DKW Inequality~\citep{DKW}]\label{Lemma: DKW Inequality}
    Given a sequence of \iid random variables $\sequence{X}{\wassmeassamplecount}$ generated from a distribution $\wassmeas$ with true cdf $\CDF_\wassmeas$. Let $\CDF_{\empwassmeas}$ be the  cdf of the empirical measure $\empwassmeas$ generated from the samples $\sequence{X}{\wassmeassamplecount}$. Then, for all $\approxerror \geq 0$, given $n \geq \frac{\ln(2)}{2\epsilon^2}$, we have:
    \begin{align*}
        \Prob\fbrac{\sup_{x \in \R} \abs{\CDF_\wassmeas(x) - \CDF_{\empwassmeas}(x)} \geq \approxerror} \leq 2\exp\fbrac{-2n\approxerror^2}
    \end{align*}
\end{lemma}


\EmpiricalMeasureBiLipschitz*

\begin{proof}
    For $\wassmeassamplecount \geq \log\fbrac{\frac{2}{\confidence}}\max\fbrac{\frac{2}{\approxerror^2\bucketsize^2\lipschitzconstant_\wassmeas^2},\frac{8\lipschitzconstant_\wassmeas^2}{\approxerror^2\bucketsize^2}}$, we have by Lemma~\ref{Lemma: DKW Inequality}:
    \begin{align*}
        \Prob\tbrac{\sup_{x\in\bucketset} \abs{\CDF_\wassmeas(x) - \CDF_{\empbucketwassmeas{\bucketsize}}(x)} \geq \min\fbrac{\frac{\approxerror\bucketsize\lipschitzconstant_\wassmeas}{2},\frac{\approxerror\bucketsize}{4\lipschitzconstant_\wassmeas}}} \leq \confidence
    \end{align*}
    Let us denote $\dkwerror = \min\fbrac{\frac{\approxerror\bucketsize\lipschitzconstant_\wassmeas}{2},\frac{\approxerror\bucketsize}{4\lipschitzconstant_\wassmeas}}$. Then, we have with probability $1 - \confidence$,
    \begin{align}
        \sup_{x\in\bucketset} \abs{\CDF_\wassmeas(x) - \CDF_{\empbucketwassmeas{\bucketsize}}(x)} \leq \dkwerror\,.\label{Eq: DKW Bound}
    \end{align} 
    Now, we have $\forall a,b \in \bucketset$, with probability $1 - \confidence$:
    \begin{align*}
        &\abs{\CDF_{\empbucketwassmeas{\bucketsize}}(a) - \CDF_{\empbucketwassmeas{\bucketsize}}(b)} \\
        \leq & \abs{\CDF_\wassmeas(a) - \CDF_\wassmeas(b)} + 2\dkwerror\\
        \leq & \lipschitzconstant_\wassmeas\abs{a - b} + \approxerror\lipschitzconstant_\wassmeas\abs{a - b}\\
        \leq & \lipschitzconstant_\wassmeas(1+\approxerror)\abs{a - b}
    \end{align*}
    Where the first inequality follows from Equation~\eqref{Eq: DKW Bound} and triangle inequality, and the second inequality follows  from the fact that $\dkwerror \leq \frac{\approxerror\bucketsize\lipschitzconstant_\wassmeas}{2}$ and $\bucketsize \leq \abs{a - b}, \forall a,b \in \bucketset$. For the other side of the inequality,
    \begin{align*}
        &\abs{\CDF_{\empbucketwassmeas{\bucketsize}}(a) - \CDF_{\empbucketwassmeas{\bucketsize}}(b)} \\
        \geq & \abs{\CDF_\wassmeas(a) - \CDF_\wassmeas(b)} - 2\dkwerror\\
        \geq & \frac{1}{\lipschitzconstant_\wassmeas}\abs{\abs{a - b}} - \frac{\approxerror\abs{a - b}}{2\lipschitzconstant_\wassmeas}\\
        \geq & \frac{1}{\lipschitzconstant_\wassmeas(1+\approxerror)}\abs{a - b}
    \end{align*}
    Where the first inequality follows from Equation~\eqref{Eq: DKW Bound} and triangle inequality, and the second inequality follows  from the fact that $\dkwerror \leq \frac{\approxerror\bucketsize}{4\lipschitzconstant_\wassmeas}$ and $\bucketsize \leq \abs{a - b}, \forall a,b \in \bucketset$, and the third inequality follows from the fact that for all $\approxerror \in (0,1)$, $1 - \frac{\approxerror}{2} \geq \frac{1}{1+\approxerror}$.
\end{proof}

%% file: Sections_PODS/Appendix/3_Distribution_Summaries.tex
\section{Estimation Guarantees of \slPDFbase{} and \slCDFbase{}}\label{Appendix: Performance Guarantee of SPA and SCA}

In this section, we provide the proofs for Theorem~\ref{Theorem: slPDFbase works} and~\ref{Theorem: slCDFbase works}. We restate the theorems here for ease of reading:

\slPDFbaseWorks*

\begin{proof}[Proof of Theorem~\ref{Theorem: slPDFbase works}]
\textbf{Proof of (a):}
    We Start with the first equation:
    \begin{align*}
       \estpdf(i) - \pdf_{\empbucketwassmeas{\bucketsize}}(i)
        = \frac{\estimatedfrequency{i}}{\max_j \estimatedcumulant{j}} - \frac{\truefrequency{i}}{\totalfrequency}
        \leq &\frac{\estimatedfrequency{i}}{\totalfrequency - 2~\residualfrequency{\countersize/4}} - \frac{\truefrequency{i}}{\totalfrequency}\\
        \leq &\frac{\truefrequency{i}}{\totalfrequency - 2~\residualfrequency{\countersize/4}} - \frac{\truefrequency{i}}{\totalfrequency}\\
        = &\truefrequency{i} \fbrac{\frac{1}{\totalfrequency - 2~\residualfrequency{\countersize/4}} - \frac{1}{\totalfrequency}}\\
        = &\frac{\truefrequency{i}}{\totalfrequency}\fbrac{\frac{2~\residualfrequency{\countersize/4}}{\totalfrequency - 2~\residualfrequency{\countersize/4}}}\\
        \leq &\frac{\truefrequency{i}}{\totalfrequency}~ \frac{4~\residualfrequency{\countersize/4}}{\totalfrequency}
    \end{align*}
    For the second part, 
    \begin{align*}
        &\pdf_{\empbucketwassmeas{\bucketsize}}(i) - \estpdf(i)
        = \frac{\truefrequency{i}}{\totalfrequency} - \frac{\estimatedfrequency{i}}{\max_j \estimatedcumulant{j}}
        \leq \frac{\truefrequency{i}}{\totalfrequency} - \frac{\estimatedfrequency{i}}{\totalfrequency}
        \leq \frac{\truefrequency{i} - \estimatedfrequency{i}}{\totalfrequency}
    \end{align*}
    Finally, we have:
    \begin{align*}
        &\abs{\pdf_{\empbucketwassmeas{\bucketsize}}(i) - \estpdf(i)}
        = \max\fbrac{\pdf_{\empbucketwassmeas{\bucketsize}}(i) - \estpdf(i),\estpdf(i) - \pdf_{\empbucketwassmeas{\bucketsize}}(i)}
        \leq \max\fbrac{\frac{4~\residualfrequency{\countersize/4}}{\totalfrequency} \frac{\truefrequency{i}}{\totalfrequency},\frac{\truefrequency{i} - \estimatedfrequency{i}}{\totalfrequency}}    
    \end{align*}
    Now, we also have by Corollary~\ref{Corollary: Mergeable Misra Gries k/4} and the fact that $\truefrequency{i}\leq \totalfrequency$,
    \begin{align*}
        \abs{\pdf_{\empbucketwassmeas{\bucketsize}}(i) - \estpdf(i)} \leq \frac{4~\residualfrequency{k/4}}{\totalfrequency}
    \end{align*}
    \textbf{Proof of (b):}
    \blue{If $\wassmeas$ is $\subgaussianparameter_\wassmeas$-sub-Gaussian, by Lemma~\ref{Theorem: Empirical Measure is Subgaussian}, we know that $\empwassmeas$ generated by $\wassmeas$ is $(\threshold,\subgaussianparameter_\wassmeas)$ sub-Gaussian given $\wassmeassamplecount \geq \constant\exp{\fbrac{\frac{\threshold^2}{\subgaussianparameter_\wassmeas^2}}}\log\fbrac{\frac{1}{\confidence}}$. Here, we can fix $\threshold = 2\subgaussianparameter_\wassmeas\sqrt{\log{\fbrac{\frac{1}{\baseapprox}}}}$.} Hence, by the property of sub-Gaussian distributions~\ref{Assumption: Bounded Tails} and the fact that each bucket of size $\bucketsize$,
    \begin{align*}
        \frac{\residualfrequency{\countersize/4}}{\totalfrequency} = \Prob_{\empwassmeas}\tbrac{X \geq \ceil*{2\subgaussianparameter_\wassmeas\sqrt{\log{\fbrac{\frac{1}{\baseapprox}}}}}} \leq \approxerror \,.
    \end{align*}
    \textbf{Proof of (c):}
    For $\weibull$-sub-Weibull distributions, by Lemma~\ref{theorem: Empirical Distribution is SubWeibull}, we know that $\empwassmeas$ generated by $\wassmeas$ is $\threshold,\weibull$-sub-Weibull given $\streamlength \geq \frac{\exp\fbrac{\sqrt[\weibull]{\threshold}}}{12}\log\frac{1}{\confidence}$. Here, we can fix $\threshold = \constant\fbrac{\log\fbrac{\frac{1}{\approxerror}}}^\weibull$, and thus $\streamlength \geq \frac{1}{3\approxerror}\log\frac{1}{\confidence}$ suffices. Hence, by the property of sub-Weibull distributions (Definition~\ref{Def: Partial sub-Weibull}) and the fact that each bucket of size $\bucketsize$,
    \begin{align*}
        \frac{\residualfrequency{\countersize/4}}{\totalfrequency} = \Prob_{\empwassmeas}\tbrac{X \geq \ceil*{\constant\fbrac{\log\fbrac{\frac{1}{\approxerror}}}^\weibull}} \leq \approxerror 
    \end{align*}
\end{proof}

\slCDFbaseWorks*

\begin{proof}[Proof of Theorem~\ref{Theorem: slCDFbase works}]

\textbf{Step 1: Bounding the residuals in frequency estimation under tail conditions.} 
    \blue{For sub-Gaussian distributions, by Lemma~\ref{Theorem: Empirical Measure is Subgaussian}, we know that $\empwassmeas$ generated by $\wassmeas$ is $(\threshold,\subgaussianparameter_\wassmeas)$ sub-Gaussian given $\wassmeassamplecount \geq \constant\exp{\fbrac{\frac{\threshold^2}{\subgaussianparameter_\wassmeas^2}}}\log\fbrac{\frac{1}{\confidence}}$. Here, we can fix $\threshold = 2\subgaussianparameter_\wassmeas\sqrt{\log{\fbrac{\frac{4}{\baseapprox}}}}$.} Hence, by the property of sub-Gaussian distributions~\ref{Assumption: Bounded Tails} and the fact that each bucket of size $\bucketsize$,
    \begin{align}
        \frac{\residualfrequency{\countersize/4}}{\totalfrequency} = \Prob_{\empwassmeas}\tbrac{X \geq \basemgcover} \leq \frac{\approxerror}{4} \label{Eq: Residual Ratio Bound}
    \end{align}
    For $\weibull$-sub-Weibull distributions, by Lemma~\ref{theorem: Empirical Distribution is SubWeibull}, we know that $\empwassmeas$ generated by $\wassmeas$ is $\fbrac{\threshold,\weibull}$-sub-Weibull given $\streamlength \geq \frac{\exp\fbrac{\sqrt[\weibull]{\threshold}}}{12}\log\frac{1}{\confidence}$. Here, we can fix $\threshold = \fbrac{\log\fbrac{\frac{4}{\approxerror}}}^\weibull$, and thus $\streamlength \geq \frac{1}{3\approxerror}\log\frac{1}{\confidence}$ suffices. Hence, by the property of sub-Weibull distributions (Definition~\ref{Def: Partial sub-Weibull}) and the fact that each bucket of size $\bucketsize$,
    \begin{align}
        \frac{\residualfrequency{\countersize/4}}{\totalfrequency} = \Prob_{\empwassmeas}\tbrac{X \geq \ceil*{\constant\fbrac{\log\fbrac{\frac{4}{\approxerror}}}^\weibull}} \leq \frac{\approxerror}{4} \label{Eq: Residual Ratio Bound Weibull 2}
    \end{align}

    \textbf{Step 2: Bounding the error in CDF estimation.} 
    
    \textit{Part i.} Now, we have:
    \begin{align*}
        \CDF_{\empbucketwassmeas{\bucketsize}}(i) - \estCDF(i) &= \frac{\truecumulant{i}}{N} - \frac{\estimatedcumulant{i}}{\max_{j}\estimatedcumulant{j}}\\
        &\leq \frac{\truecumulant{i}-\estimatedcumulant{i}}{\totalfrequency} &\text{since }\totalfrequency \geq \max_j \estimatedcumulant{j}\\
        &=\frac{2~\residualfrequency{\countersize/4}}{\totalfrequency} &\text{By Lemma~\ref{Lemma: Estimate Cumulant Works}}\\
        &\leq \baseapprox &\text{By Equation~\eqref{Eq: Residual Ratio Bound} and~\eqref{Eq: Residual Ratio Bound Weibull 2}}
    \end{align*}
    \textit{Part ii.}  Now, we look into the other side of the error. First, we observe that:
    \begin{align*}
        \frac{\residualfrequency{\countersize/4}}{\totalfrequency - 2~\residualfrequency{\countersize/4}} \leq \frac{2~\residualfrequency{\countersize/4}}{\totalfrequency} \leq \approxerror
    \end{align*}
    Here, the first inequality is given by the fact $2\residualfrequency{\countersize/4} \leq \totalfrequency$, and the second inequality follows from~\eqref{Eq: Residual Ratio Bound}.

Thus, 
    \begin{align*}
        \estCDF(i) - \CDF_{\empbucketwassmeas{\bucketsize}}(i) &= \frac{\estimatedcumulant{i}}{\max_{j}\estimatedcumulant{j}} - \frac{\truecumulant{i}}{N}\\
        &\leq \frac{\truecumulant{i}}{\max_{j}\estimatedcumulant{j}} - \frac{\truecumulant{i}}{N}\,, &\text{since }\truecumulant{i} \geq \estimatedcumulant{i}\\
        &=\truecumulant{i}\fbrac{\frac{1}{\max_j \estimatedcumulant{j}} - \frac{1}{\totalfrequency} }\\
        &\leq \truecumulant{i}\fbrac{\frac{1}{\totalfrequency - 2~\residualfrequency{\countersize/4}} - \frac{1} {\totalfrequency} }\,, &\text{since }\max_j \estimatedcumulant{j} \geq \max_j \truecumulant{j} - 2~\residualfrequency{\countersize/4} = \totalfrequency - \residualfrequency{\countersize/4}\\
        &=\truecumulant{i}\fbrac{\frac{2~\residualfrequency{\countersize/4}}{\totalfrequency\fbrac{\totalfrequency-2~\residualfrequency{\countersize/4}}}}\\
        &\leq \frac{2~\residualfrequency{\countersize/4}}{\totalfrequency - 2~\residualfrequency{\countersize/4}} &\truecumulant{i} \leq \totalfrequency\\
        &\leq \frac{4~\residualfrequency{\countersize/4}}{\totalfrequency} &\totalfrequency \geq 4~\residualfrequency{\countersize/4}\\
        &\leq \approxerror &\text{By Equation~\eqref{Eq: Residual Ratio Bound}~and~\eqref{Eq: Residual Ratio Bound Weibull 2}}
    \end{align*}

\end{proof}

%% file: Sections_PODS/Appendix/4_Wasserstein_Approximation.tex
\section{Estimation Guarantee of \slwassbase}\label{Appendix: SWA Guarantees}

In this section, we state the proofs for Theorem~\ref{Theorem: slWassBase Works}. We start with Lemma~\ref{Lemma: Bucketing Wasserstein Approximation} which shows that a bucketed empirical measure $\empbucketwassmeas{\bucketsize}$ is close to the underlying empirical measure $\empwassmeas$ in Wasserstein distance given the bucket size is sufficiently small.

\begin{lemma}\label{Lemma: Bucketing Wasserstein Approximation}
    Given a distribution $\wassmeas$ and corresponding $\bucketdisc{\wassmeas}{\bucketsize}$ 
    with bucket size $\bucketsize$, we have:
    \begin{align*}
        \Wasserstein{p}{\wassmeas}{\bucketdisc{\wassmeas}{\bucketsize}} \leq \bucketsize
    \end{align*}
\end{lemma}
\begin{proof}
For any $i$-th bucket corresponding to $\bucketdisc{\wassmeas}{\bucketsize}$, we have:
\begin{align*}
    \CDF_{\wassmeas}(\bucketset_i + \frac{\bucketsize}{2}) = \CDF_{\bucketdisc{\wassmeas}{\bucketsize}}(\bucketset_i + \frac{\bucketsize}{2})
\end{align*}

    \begin{align*}
        \abs{\invCDF_{\wassmeas}(x) - \invCDF_{\bucketdisc{\wassmeas}{\bucketsize}}(x)} \leq \bucketsize
    \end{align*}
    Hence, we have:
    \begin{align*}
        &\Wasserstein{p}{\wassmeas}{\bucketdisc{\wassmeas}{\bucketsize}}
        =\fbrac{\int_0^1 \abs{\invCDF_{\wassmeas}(x) - \invCDF_{\bucketdisc{\wassmeas}{\bucketsize}}(x)}^p \,dx}^{\frac{1}{p}}
        \leq \fbrac{\int_0^1 \bucketsize^p \,dx}^{\frac{1}{p}}
        = \bucketsize 
    \end{align*}
    This concludes the proof.
\end{proof}
A general version of Lemma~\ref{Lemma: Bucketing Wasserstein Approximation} is given in~\citet[Theorem 4.1]{DBLP:conf/nips/StaibCSJ17}. However, we prove our version here for simplicity and ease of access.

Next, we prove the following lemma that characterizes the approximation of a function based on an approximation guarantee on its inverse.

\begin{restatable}[Inverse Approximation of inverse $\lipschitzconstant$-\lipschitz functions]{lemma}{inverseapproximation}\label{Lemma: Approximation of Inverse Lipschitz Functions}
    Let $f$ be a $\lipschitzconstant$ bi-\lipschitz function. Given an estimate $\fhat$ of $f$ such that $\fhat(x) \in [f(x)-\approxerror,f(x)+\approxerror]$, we can construct an estimate $\invfhat$ of the pseudoinverse function $\inverse{f}$ such that $\forall x \in \range(f)$:
    \begin{align*}
        \invfhat(x) \in [\inverse{f}(x)-\lipschitzconstant\approxerror,\inverse{f}(x)+\lipschitzconstant\approxerror]
    \end{align*}
\end{restatable}

\begin{proof}
    We construct $\invfhat$ as $\invfhat(y) = \min_x\sbrac{x \in \R\cup\sbrac{-\infty}:\fhat(x) = y}$. Let us consider $x \in \range(f)$, and $\inverse{f}(x) = y_1$ and $\invfhat(x) = y_2$, and w.l.o.g. assume $y_2 \geq y_1$. Then, we have:
    \begin{align*}
        f(y_1) &= \fhat(y_2)\\
        \fhat(y_1) + \approxerror &\geq \fhat(y_2)\\
        \approxerror &\geq \fhat(y_2) - \fhat(y_1)
    \end{align*}
    Next, we use the fact that the function $f$ is $\lipschitzconstant$ bi-\lipschitz to show that $y_2 - y_1$ is bounded by $\lipschitzconstant\approxerror$:
    \begin{align*}
        &y_2 - y_1\\
        \leq &\invfhat(\fhat(y_2)) - \invfhat(\fhat(y_1)) &\text{By definition of $\invfhat$}\\
        \leq &\lipschitzconstant(\fhat(y_2)-\fhat(y_1))
        \leq \lipschitzconstant\approxerror
    \end{align*}
    Hence, for any $x\in \range(f)$, we have $|\invfhat(x) - \inverse{f}(x)| \leq \lipschitzconstant\approxerror$, completing our proof.
\end{proof}

The following corollary follows directly.

\begin{corollary}[From CDF to Inverse CDF]\label{Corollary: Inverse CDF Approximation}
    For an $\lipschitzconstant$ bi-Lipschitz distribution $\wassmeas$ with CDF $\CDF_\wassmeas$ and pseudoinverse CDF $\invCDF_\wassmeas$, given an approximate CDF $\wh{\CDF}_\wassmeas$ satisfying $|\wh{\CDF}_\wassmeas(x) - \CDF_\wassmeas(x)| \leq \approxerror, \forall x \in \R$, we construct an approximation of pseudoinverse CDF $\wh{\invCDF_\wassmeas}$, such that $\forall x \in [0,1]$, $|\wh{\invCDF_\wassmeas}(x) - \invCDF_\wassmeas(x)| \leq \lipschitzconstant\approxerror$.
\end{corollary}

Now, we prove the Theorem~\ref{Theorem: slWassBase Works}. We restate the theorems here for ease of reading:

\slWassBaseWorks*

\begin{proof}[Proof of Theorem~\ref{Theorem: slWassBase Works}]
    By Theorem~\ref{Theorem: Empirical Measure is Bi-Lipschitz},~\ref{Theorem: slCDFbase works}, and Lemma~\ref{Lemma: Approximation of Inverse Lipschitz Functions}; and the parameters we have used, we have with probability $1 - \confidence$:
    \begin{align*}
        \abs{\invCDF_{\empbucketwassmeas{\approxerror/2}}-\estinvCDF} \leq 2\lipschitzconstant_\wassmeas\abs{\CDF_{\empbucketwassmeas{\approxerror/2}}  - \estCDF} \leq \approxerror/2
    \end{align*}
    Hence, by Lemma~\ref{Lemma: Univariate Wasserstein Characterization}, we have with probability at least $1-\confidence$:
    \begin{align*}
        \Wasserstein{p}{\empwassmeas}{\estwassmeas} &\leq \Wasserstein{p}{\empwassmeas}{\empbucketwassmeas{\approxerror/2}} + \Wasserstein{p}{\empbucketwassmeas{\approxerror/2}}{\estwassmeas}\\
        &\leq \approxerror/2 + \fbrac{\int_0^1 |\invCDF_{\empbucketwassmeas{\approxerror/2}}(r)-\estinvCDF(r)|^p\,dr}^{1/p}\\
        &\leq \approxerror/2 + \fbrac{\int_0^1 |\approxerror/2|^p\,dr}^{1/p} = \approxerror
    \end{align*}

    The bound on number of buckets can be obtained by plugging in the values in the line~\ref{SWA Line: SCA Call} of Algorithm~\ref{Algorithm: Base Wasserstein Approximation Algorithm} in Theorem~\ref{Theorem: slCDFbase works}.
\end{proof}

We now proof the Theorem~\ref{Theorem: SWA Estimation Guarantee w.r.t. the True Measure}, restated here for ease of reading:

\slwassbasePAC*

\begin{proof}
    We combine the results of Corollary~\ref{Corollary: Concentration in Wasserstein distance of 1-D Empirical Measures Subgaussian}~,~\ref{Corollary: Concentration in Wasserstein distance of 1-D Empirical Measures Subweibull}, and Theorem~\ref{Theorem: slWassBase Works} to establish this result. By Corollary~\ref{Corollary: Concentration in Wasserstein distance of 1-D Empirical Measures Subgaussian} for sub-Gaussian distributions, we have with probability $1-2\confidence$:
    \begin{align}
        \abs{\Wasserstein{1}{\wassmeas}{\wassmeastwo} - \Wasserstein{1}{\empwassmeas}{\empwassmeastwo}} &\leq 2\approxerror &\text{By Triangle Inequality}\label{Equation: SWA PAC 1}
    \end{align}
    Similarly, by Corollary~\ref{Corollary: Concentration in Wasserstein distance of 1-D Empirical Measures Subweibull} for $\weibull$-sub-Weibull distributions, we have with probability $1-2\confidence$:
    \begin{align}
        \abs{\Wasserstein{1}{\wassmeas}{\wassmeastwo} - \Wasserstein{1}{\empwassmeas}{\empwassmeastwo}} &\leq 2\approxerror &\text{By Triangle Inequality}\label{Equation: SWA PAC 1 SubW}
    \end{align}
    By Theorem~\ref{Theorem: SWA Estimation Guarantee w.r.t. the True Measure}, we have with probability $1-2\confidence$:
    \begin{align}
        \abs{\Wasserstein{1}{\estwassmeas}{\estwassmeastwo} - \Wasserstein{1}{\empwassmeas}{\empwassmeastwo}} &\leq 2\approxerror &\text{By Triangle Inequality}\label{Equation: SWA PAC 2}
    \end{align}
    A union bound argument and triangle inequality over Equations~\eqref{Equation: SWA PAC 1}~or~\eqref{Equation: SWA PAC 1 SubW} and~\eqref{Equation: SWA PAC 2} completes the proof.
\end{proof}

%% file: Sections_PODS/Appendix/4.5_TV_Distance_Estimation.tex
\section{Estimation Guarantee of \slTVbase{}}\label{Appendix: Proof of STVA}

In this section, we prove the theorems concerning \slTVbase{}, i.e. Theorem~\ref{Theorem: bucket width for Subgaussian TV}, Lemma~\ref{lemma: Conc of TV distance}, and Theorems~\ref{Theorem: Tv Distance Approx Algo Works}~and~\ref{Theorem: PAC of SLTV}.

\subsection{Proof of Theorem~\ref{Theorem: bucket width for Subgaussian TV}}\label{Appendix: Proof for bucket width of STVA}

In this section, we detail the proof of Theorem~\ref{Theorem: bucket width for Subgaussian TV}.

\bucketingtv*


\begin{proof}
    \textbf{Initial Bounds: }The distribution $\wassmeas$ has a $\lipschitzconstant_\wassmeas$-\lipschitz PDF. Consider any bucket $\bucketelement$ of size $\bucketsize$, then $\min_{y \in \bucketelement} \wassmeas(y) - \max_{y \in \bucketelement} \wassmeas(y) \leq \bucketsize\lipschitzconstant_\wassmeas$. As $\min_{y \in \bucketelement} \wassmeas(y) \leq \bucketwassmeas{\bucketsize}(x) \leq \max_{y \in \bucketelement} \wassmeas(y), \forall x \in \bucketelement $, we have for any bucket $\bucketelement$:
    \begin{align}
        \nonumber \abs{\wassmeas(x) - \bucketwassmeas{\bucketsize}(x)} &\leq \bucketsize\lipschitzconstant_\wassmeas &\forall x \in \bucketelement\\
        \int_\bucketelement \abs{\wassmeas(x) - \bucketwassmeas{\bucketsize}(x)} \dd x &\leq \bucketsize^2\lipschitzconstant_\wassmeas\label{Equation: Intra Bucket Bound Subweibull}
    \end{align}

    The main idea of our proof is that we bound the error due to bucketing in buckets with high frequency and use the fact that the remaining buckets has sufficiently small frequency to achieve our bound. To denote the buckets with high frequency, recall that $\topk{k}$ denotes the set of $k$ buckets with highest frequency, $\bucketsize$ denotes the length of each bucket.

    \begin{align}
        \nonumber& \TV{\wassmeas}{\bucketwassmeas{\bucketsize}}\\
        =& \frac{1}{2}\int_{-\infty}^{\infty} \abs{\wassmeas(x) - \bucketwassmeas{\bucketsize}(x)} \dd x\\\nonumber
        =& \frac{1}{2} \sum_{\bucketelement \in \bucketset} \int_\bucketelement \abs{\wassmeas(x) - \bucketwassmeas{\bucketsize}(x)} \dd x\\\nonumber
        =& \frac{1}{2} \sum_{\bucketelement \in \topk{k}} \int_\bucketelement \abs{\wassmeas(x) - \bucketwassmeas{\bucketsize}(x)} \dd x + \frac{1}{2} \sum_{\bucketelement \notin \topk{k}} \int_\bucketelement \abs{\wassmeas(x) - \bucketwassmeas{\bucketsize}(x)} \dd x\\\nonumber
        \leq& \frac{1}{2} \sum_{\bucketelement \in \topk{k}}  \bucketsize^2\lipschitzconstant_\wassmeas + \frac{1}{2} \sum_{\bucketelement \notin \topk{k}} \tbrac{\int_\bucketelement \bucketwassmeas{\bucketsize}(x) \dd x + \int_\bucketelement \wassmeas(x) \dd x}&\text{By Equation~\eqref{Equation: Intra Bucket Bound Subweibull}}\\\nonumber
        \leq& k\bucketsize^2\lipschitzconstant_\wassmeas/2 +  \sum_{\bucketelement \notin \topk{k}}  \int_\bucketelement \wassmeas(x) \dd x\label{Equation: TV Base Bound}\\
    \end{align}
    \textbf{Sub-Weibull Distributions: } For $\weibull$-sub-Weibull distributions, we fix $k = \frac{\lipschitzconstant_\wassmeas\fbrac{\log(2\constant_\weibull/\approxerror)}^{2\weibull}}{\approxerror}$ and bucket size $\bucketsize = \frac{\approxerror}{\lipschitzconstant_\wassmeas\fbrac{\log\fbrac{2\constant_\weibull/\approxerror}}^{\weibull}}$ and bound the terms in Equation~\eqref{Equation: TV Base Bound}. For the first term, we have:

    \begin{align}
        k\bucketsize^2\frac{\lipschitzconstant_\wassmeas}{2}
        =  \frac{\lipschitzconstant_\wassmeas\fbrac{\log(2\constant_\weibull/\approxerror)}^{2\weibull}}{\approxerror}  \fbrac{\frac{\approxerror}{\lipschitzconstant_\wassmeas\fbrac{\log\fbrac{2\constant_\weibull/\approxerror}}^{\weibull}}}^2  \frac{\lipschitzconstant_\wassmeas}{2}
        \leq  \approxerror/2\label{Equation: TV Bound 1}
    \end{align}

    For the second term:
    \begin{align}\nonumber
        \sum_{\bucketelement \notin \topk{\frac{\lipschitzconstant_\wassmeas\fbrac{\log(2/\approxerror)}^{2\weibull}}{\approxerror}}} \int_\bucketelement \wassmeas(x) \dd x
        =& \Prob_{X \sim \wassmeas}\tbrac{X \in \cup_{\bucketelement \notin \topk{\frac{\lipschitzconstant_\wassmeas\fbrac{\log(2/\approxerror)}^{2\weibull}}{\approxerror}}} \bucketelement}\\
        \leq& \Prob_{X \sim \wassmeas}\tbrac{X \geq \fbrac{\log\frac{2\constant_\weibull}{\approxerror}}^\weibull}
        \leq \approxerror/2\label{Equation: TV Bound 2}
    \end{align}
    Combining Equations~\eqref{Equation: TV Bound 1} and~\eqref{Equation: TV Bound 2} with the bound from Equation~\eqref{Equation: TV Base Bound}, we obtain the result.

\textbf{Sub-Gaussian Distributions: }
    For sub-Gaussian distributions, we fix $k = \frac{\subgaussianparameter_\wassmeas^2\lipschitzconstant_\wassmeas\log(2/\approxerror)}{\approxerror}$ and bucket size $\bucketsize = \frac{\approxerror}{\subgaussianparameter_\wassmeas\lipschitzconstant_\wassmeas\sqrt{\log\fbrac{2/\approxerror}}}$ and bound the terms in Equation~\eqref{Equation: TV Base Bound}. For the first term, we have

    \begin{align}
        k\bucketsize^2\frac{\lipschitzconstant_\wassmeas}{2}
        =  \frac{\subgaussianparameter_\wassmeas^2\lipschitzconstant_\wassmeas\log(2/\approxerror)}{\approxerror}  \fbrac{\frac{\approxerror}{\subgaussianparameter_\wassmeas\lipschitzconstant_\wassmeas\sqrt{\log\fbrac{2/\approxerror}}}}^2  \frac{\lipschitzconstant_\wassmeas}{2}
        \leq  \approxerror/2\label{Equation: TV Bound 1 SubW}
    \end{align}

    For the second term:
    \begin{align}\nonumber
        \sum_{\bucketelement \notin \topk{\frac{\subgaussianparameter_\wassmeas^2\lipschitzconstant_\wassmeas\log^2(2/\approxerror)}{\approxerror}}} \int_\bucketelement \wassmeas(x) \dd x
        =& \Prob_{X \sim \wassmeas}\tbrac{X \in \cup_{\bucketelement \notin \topk{\frac{\subgaussianparameter_\wassmeas^2\lipschitzconstant_\wassmeas\log(2/\approxerror)}{\approxerror}}} \bucketelement}\\\nonumber
        \leq& \Prob_{X \sim \wassmeas}\tbrac{X \geq \subgaussianparameter_\wassmeas\sqrt{\log\fbrac{2/\approxerror}}}\\
        \leq& \approxerror/2\label{Equation: TV Bound 2 SubW}
    \end{align}
    Combining Equations~\eqref{Equation: TV Bound 1 SubW} and~\eqref{Equation: TV Bound 2 SubW} with the bound from Equation~\eqref{Equation: TV Base Bound}, we obtain the result.
\end{proof}

\subsection{Proof of Lemma~\ref{lemma: Conc of TV distance}}
 
 Consider the two possible ways of looking at a histogram. Given a histogram defined over $\R$, one can think of it as defining a (not necessarily probability) measure over $\R$ defined so that the measure at each point is equal to the frequency of the bucket the point lies in. Alternatively, one can define a measure over the buckets with the measure at each bucket is equal to the frequency of the bucket. In this section, we formalize these notions and establish the relation between them in term of TV and Wasserstein distances.

\begin{definition}[Bucketed Continuous Distribution]\label{Definition: Bucketed Continuous Distribution}
    Let $\distribution$ be a distribution supported on a (finite or infinite) closed interval $I \subseteq \R$ with (discrete or continuous) measure $\measure$. Given $x_0$ as a reference point in $I$ and $\bucketsize$ as the bucket width, we further represent $I$ as $I(x_0, J, \bucketsize) \triangleq \cup_{j \in J} [x_0 + j\bucketsize, x_0 + (j+1)\bucketsize]$. Here, index set $J \subseteq \mathbb{Z}$. We define the corresponding bucketed continuous distribution $\bucketcont{\distribution}{\bucketsize}$ with measure $\bucketcont{\measure}{\bucketsize}$ defined over $I \subseteq \R$ as:
    \begin{align*}
        \bucketcont{\measure}{\bucketsize}(x) &= \frac{1}{\bucketsize}\int_{x_0 + j\bucketsize}^{x_0 + (j+1)\bucketsize} \measure(x)\dd x\\
        &= \frac{1}{\bucketsize}\measure\fbrac{[x_0 + j\bucketsize, x_0 + (j+1)\bucketsize]}\,,
    \end{align*}
\end{definition}
where $x \in [x_0 + j\bucketsize, x_0 + (j+1)\bucketsize]$.
\begin{definition}[Bucketed Discrete Distribution]\label{Definition: Bucketed Discrete Distribution}
    Let $\distribution$ be a distribution supported on a (finite or infinite) closed interval $I \subseteq \R$ with (discrete or continuous) measure $\measure$. Given $x_0$ as a reference point in $I$ and $\bucketsize$ as the bucket width, we further represent $I$ as $I(x_0, J, \bucketsize) \triangleq \cup_{j \in J} [x_0 + j\bucketsize, x_0 + (j+1)\bucketsize]$. We define the corresponding bucketed discrete distribution $\bucketdisc{\distribution}{\bucketsize}$ with measure $\bucketdisc{\measure}{\bucketsize}$ defined over $\cup_{j \in J} \{x_0 + j\bucketsize + b/2\} \subseteq \R$ as $\bucketdisc{\measure}{\bucketsize}(j) =  \int_{x_0 + j\bucketsize}^{x_0 + (j+1)\bucketsize} \measure(x)\dd x = \measure\fbrac{[x_0 + j\bucketsize, x_0 + (j+1)\bucketsize]}$.
\end{definition}
The support set of $\bucketdisc{\measure}{\bucketsize}$ inherits the metric structure of $\R$. For ease of notation, we express the distance between the $i$-th and $j$-th support points of $\bucketdisc{\measure}{\bucketsize}$ as $d(\bucketelement_i,\bucketelement_j) \triangleq |i-j|b$.

We introduce the following lemma establishing the equivalence of bucketed discrete and continuous measure with respect to the TV distance.

\begin{restatable}[TV of continuous and discrete bucketed distributions]{lemma}{tvcontinuousdiscrete}\label{Lemma: tv_cont_to_discrete}
    Given two measures $\wassmeas$, and $\wassmeastwo$, we have:
    \begin{align*}
        \TV{\bucketcont{\wassmeas}{\bucketsize}}{\bucketcont{\wassmeastwo}{\bucketsize}} = \TV{\bucketdisc{\wassmeas}{\bucketsize}}{\bucketdisc{\wassmeastwo}{\bucketsize}} 
    \end{align*}
\end{restatable}

\begin{proof}
    \begin{align*}
        \TV{\bucketcont{\wassmeas}{\bucketsize}}{\bucketcont{\wassmeastwo}{\bucketsize}}
        = & \frac{1}{2}\int \abs{\bucketcont{\wassmeas}{\bucketsize}(x) - \bucketcont{\wassmeastwo}{\bucketsize}(x)} \dd x\\
        = & \frac{1}{2} \sum_{j \in J} \frac{1}{\bucketsize} \int_{x_0 + j\bucketsize}^{x_0 + (j+1)\bucketsize} \abs{\int_{x_0 + j\bucketsize}^{x_0 + (j+1)\bucketsize} \wassmeas(x)\dd x - \int_{x_0 + j\bucketsize}^{x_0 + (j+1)\bucketsize} \wassmeastwo(x)\dd x}\dd x\\
        = & \frac{1}{2} \sum_{j \in J} \abs{\int_{x_0 + j\bucketsize}^{x_0 + (j+1)\bucketsize} \wassmeas(x)\dd x - \int_{x_0 + j\bucketsize}^{x_0 + (j+1)\bucketsize} \wassmeastwo(x)\dd x}\\
        = & \TV{\bucketdisc{\wassmeas}{\bucketsize}}{\bucketdisc{\wassmeastwo}{\bucketsize}}
    \end{align*}
\end{proof}

We now establish a TV distance concentration result on bucketed empirical distribution corresponding to a sub-Gaussian distribution.


\TVdistanceconc*

\begin{proof}
    Let us consider the standard bijection from $\field{Z}$ to $\Nat$ as:
    \begin{align*}
    f(x) =
        \begin{dcases*}
            $2x$ & if $x > 0$ \\ 
            $-2x+1$ & if $x \leq 0$ 
        \end{dcases*}
    \end{align*}

\textbf{$\subgaussianparameter_\wassmeas$-Sub-Gaussian Distributions: }
    For sub-Gaussian distributions,If $i$ is even, we have:
    \begin{align*}
        \pdf(i) &= \bucketdisc{\wassmeas}{\bucketsize}(\bucketelement_{i/2})
        \leq \bucketdisc{\wassmeas}{\bucketsize}\fbrac{\cup_{j \geq i/2}\bucketelement_{j}}
        \leq \wassmeas(\sbrac{x|x\geq \bucketsize i/2})
        \leq 2\exp\fbrac{-\frac{i^2\bucketsize^2}{4\subgaussianparameter_\wassmeas^2}}\,.
    \end{align*}

    If $i$ is odd, we have:

    \begin{align*}
        \pdf(i) &= \bucketdisc{\wassmeas}{\bucketsize}(\bucketelement_{-(i-1)/2})
        \leq \bucketdisc{\wassmeas}{\bucketsize}\fbrac{\cup_{j \leq -(i-1)/2}\bucketelement_{j}}
        \leq \wassmeas(\sbrac{x|x\leq -\bucketsize (i-1)/2})
        \leq 2\exp\fbrac{-\frac{(i-1)^2\bucketsize^2}{4\subgaussianparameter_\wassmeas^2}}\,.
    \end{align*}
    Combining these terms, we have:
     \begin{align*}
        \sum_{i \in \Nat} \sqrt{\pdf_\wassmeas(i)} &\leq \sum_{i \in \Nat} 4\exp\fbrac{-i^2\bucketsize^2 /4\subgaussianparameter_\wassmeas^2}
        \leq 4 \int_{0}^{\infty} \exp(-x^2\bucketsize^2 /\subgaussianparameter_\wassmeas^2) \dd x
        = 2\sqrt{4\pi\subgaussianparameter_\wassmeas^2/\bucketsize^2}
        = \frac{4\subgaussianparameter_\wassmeas\sqrt{\pi}}{\bucketsize}\,.
    \end{align*}

    We now combine this with Lemma~\ref{Lemma: Infinite TV True Measure Concentration} to complete the proof.
    
    \textbf{$\weibull$-SubWeibull Distributions: } For $\weibull$-SubWeibull distributions, if $i$ is even, we have:
    \begin{align*}
        \pdf(i) &= \bucketdisc{\wassmeas}{\bucketsize}(\bucketelement_{i/2})
        \leq \bucketdisc{\wassmeas}{\bucketsize}\fbrac{\cup_{j \geq i/2}\bucketelement_{j}}
        \leq \wassmeas(\sbrac{x|x\geq \bucketsize i/2})
        \leq \constant_\weibull\exp\fbrac{-(i \bucketsize)^{1/\weibull}}\,.
    \end{align*}

    If $i$ is odd, we have:

    \begin{align*}
        \pdf(i) &= \bucketdisc{\wassmeas}{\bucketsize}(\bucketelement_{-(i-1)/2})
        \leq \bucketdisc{\wassmeas}{\bucketsize}\fbrac{\cup_{j \leq -(i-1)/2}\bucketelement_{j}}
        \leq \wassmeas(\sbrac{x|x\leq -\bucketsize (i-1)/2})
        \leq \constant_\weibull\exp\fbrac{-((i-1) \bucketsize)^{1/\weibull}}\,.
    \end{align*}
    Combining these terms, we have:
     \begin{align*}
        \sum_{i \in \Nat} \sqrt{\pdf_\wassmeas(i)} &\leq \sum_{i \in \Nat} 2\constant_\weibull\exp\fbrac{-(i \bucketsize)^{1/\weibull}}
        \leq 2\constant_\weibull \int_{0}^{\infty} \exp\fbrac{-(x \bucketsize)^{1/\weibull}} \dd x
        = 2\constant_\weibull \frac{\Gamma(1+\alpha)}{b}\,.
    \end{align*}
    We now combine this with Lemma~\ref{Lemma: Infinite TV True Measure Concentration} to complete the proof.
\end{proof}

%% file: Sections_PODS/Appendix/5_Bucketing_TV.tex
\subsection{Bounding Errors of \slPDFbase{} and \slTVbase{} (Theorems~\ref{Theorem: Tv Distance Approx Algo Works}~and~\ref{Theorem: PAC of SLTV})}

\slTVbaseWorks*

\begin{proof}
\textbf{Step 1: Bounding the residuals in frequency estimation under tail conditions.}
    \blue{For sub-Gaussian distributions, by Theorem~\ref{Theorem: Empirical Measure is Subgaussian}, we know that $\empbucketwassmeas{\bucketsize}$ generated by $\bucketdisc{\wassmeas}{\bucketsize}$ is $(\threshold,\subgaussianparameter_\wassmeas)$ sub-Gaussian given $\wassmeassamplecount \geq \constant\exp{\fbrac{\frac{\threshold^2}{\subgaussianparameter_\wassmeas^2}}}\log\fbrac{\frac{1}{\confidence}}$. Here, we can fix $\threshold = 2\subgaussianparameter_\wassmeas\sqrt{\log{\fbrac{\frac{6}{\baseapprox}}}}$.} Hence, by the property of sub-Gaussian distributions~\ref{Assumption: Bounded Tails} and the fact that length of each bucket is $\bucketsize$,
    \begin{align}
        \frac{\residualfrequency{\countersize/4}}{\totalfrequency} \leq \Prob_{\empbucketwassmeas{\bucketsize}}\tbrac{X \geq \basemgcoverpdf} \leq \frac{\approxerror}{6} \label{Eq: Residual Ratio Bound 2}
    \end{align}

    For $\weibull$-sub-Weibull distributions, by Lemma~\ref{theorem: Empirical Distribution is SubWeibull}, we know that $\empwassmeas$ generated by $\wassmeas$ is $\fbrac{\threshold,\weibull}$-sub-Weibull given $\streamlength \geq \frac{\exp\fbrac{\sqrt[\weibull]{\threshold}}}{12}\log\frac{1}{\confidence}$. Here, we can fix $\threshold = 2\fbrac{\log\fbrac{\frac{6}{\approxerror}}}^\weibull$, and thus $\streamlength \geq \frac{1}{2\approxerror}\log\frac{1}{\confidence}$ suffices. Hence, by the property of sub-Weibull distributions (Definition~\ref{Def: Partial sub-Weibull}) and the fact that each bucket of size $\bucketsize$,
    \begin{align}
        \frac{\residualfrequency{\countersize/4}}{\totalfrequency} = \Prob_{\empwassmeas}\tbrac{X \geq \ceil*{\constant\fbrac{\log\fbrac{\frac{6}{\approxerror}}}^\weibull}} \leq \frac{\approxerror}{4} \label{Eq: Residual Ratio Bound 2 Weibull}
    \end{align}

\textbf{Step 2: Bounding the error in PDF estimation.}
    Now, we proceed to bound the TV Distance. We denote by $\topk{\countersize}$ the set of $\countersize$ elements with highest true frequency $\truefrequency{i}$. Then, we have:
    \begin{align*}
        \norm{\estpdf - \pdf_{\empbucketwassmeas{\bucketsize}}}_1& = \sum_{i} \abs{\pdf_{\empbucketwassmeas{\bucketsize}}(i) - \estpdf(i)}\\
        & \leq \sum_{i} \max\fbrac{\frac{4~\residualfrequency{\countersize/4}}{\totalfrequency}\cdot\frac{\truefrequency{i}}{\totalfrequency},\frac{\truefrequency{i} - \estimatedfrequency{i}}{\totalfrequency}}&\text{Theorem~\ref{Theorem: slPDFbase works}}\\
        & \leq \sum_i \frac{4~\residualfrequency{\countersize/4}}{\totalfrequency}\cdot\frac{\truefrequency{i}}{\totalfrequency} + \sum_i \frac{\truefrequency{i} - \estimatedfrequency{i}}{\totalfrequency}\\
        & \leq \frac{4~\residualfrequency{\countersize/4}}{\totalfrequency}\sum_i \frac{\truefrequency{i}}{\totalfrequency} + \sum_{i \in \topk{\countersize/4}} \frac{\truefrequency{i}-\estimatedfrequency{i}}{\totalfrequency} + \sum_{i \notin \topk{\countersize/4}} \frac{\truefrequency{i}}{\totalfrequency}\\
        & \leq \frac{4~\residualfrequency{\countersize/4}}{\totalfrequency} + \frac{\sum_{i \in Learning\topk{\countersize/4}} \truefrequency{i} - \estimatedfrequency{i} + \sum_{i \notin \topk{\countersize/4}} \truefrequency{i}}{\totalfrequency}\\
        & \leq \frac{4~\residualfrequency{\countersize/4}}{\totalfrequency} + \frac{2~\residualfrequency{\countersize/4}}{\totalfrequency}&\text{Lemma~\ref{Theorem: Mergeable Misra Gries}}\\
        & = \frac{6~\residualfrequency{\countersize/4}}{\totalfrequency}\\
        & \leq \approxerror &\text{Equation~\eqref{Eq: Residual Ratio Bound 2}~and~\eqref{Eq: Residual Ratio Bound 2 Weibull}}
    \end{align*}
\end{proof}

We now proof the Theorem~\ref{Theorem: PAC of SLTV}, restated here for ease of reading:

\slTVbasePAC*

\begin{proof}
    Composing Lemma~\ref{lemma: Conc of TV distance}, Lemma~\ref{Lemma: tv_cont_to_discrete}, Theorem ~\ref{Theorem: bucket width for Subgaussian TV}, and Theorem~\ref{Theorem: Tv Distance Approx Algo Works} yields the proof. 

    By Theorem~\ref{Theorem: bucket width for Subgaussian TV}, and Lemma~\ref{Lemma: tv_cont_to_discrete}, we have:
    \begin{align}
        \nonumber\abs{\TV{\wassmeas}{\wassmeastwo} - \TV{\bucketcont{\wassmeas}{\bucketsize}}{\bucketcont{\wassmeastwo}{\bucketsize}}} &\leq  2\approxerror &\text{By Theorem~\ref{Theorem: bucket width for Subgaussian TV} and Triangle Inequality}\\
        \abs{\TV{\wassmeas}{\wassmeastwo} - \TV{\bucketdisc{\wassmeas}{\bucketsize}}{\bucketdisc{\wassmeastwo}{\bucketsize}}} &\leq  2\approxerror &\text{By Lemma~\ref{Lemma: tv_cont_to_discrete}}\label{Equation: SLTV PAC 1}
    \end{align}
    From Lemma~\ref{lemma: Conc of TV distance}, and fixing the $\bucketsize$, we have with probability $1-2\confidence$:
    \begin{align}
        \abs{\TV{\empbucketwassmeas{\bucketsize}}{\empbucketwassmeastwo{\bucketsize}} - \TV{\bucketdisc{\wassmeas}{\bucketsize}}{\bucketdisc{\wassmeastwo}{\bucketsize}}} &\leq 2\approxerror &\text{By Triangle Inequality}\label{Equation: SLTV PAC 2}
    \end{align}
    From Theorem~\ref{Theorem: Tv Distance Approx Algo Works}, we have with probability $1-2\confidence$:
    \begin{align}
        \abs{\TV{\estwassmeas}{\estwassmeastwo} - \TV{\empbucketwassmeas{\bucketsize}}{\empbucketwassmeastwo{\bucketsize}}} &\leq 2\approxerror &\text{By Triangle Inequality}\label{Equation: SLTV PAC 3}
    \end{align}
    A union bound argument and triangle inequality over Equation~\eqref{Equation: SLTV PAC 1},~\eqref{Equation: SLTV PAC 2} and~\eqref{Equation: SLTV PAC 3} completes the proof.
\end{proof}

%% file: Sections_PODS/Appendix/6_applications.tex




%% file: Sections_PODS/Appendix/9_Empirical_Tail_Bounds.tex

%% file: Sections_PODS/Appendix/7_useful_results.tex
\section{Useful Technical Results}\label{Appendix: Useful Technical Results}

\begin{lemma}[Inverse of Bi-\lipschitz function is Bi-\lipschitz]\label{Lemma: Inverse of BiLipschitz function is BiLipschitz}
	Given an invertible function $\func{f}{\dom{f}}{\range{f}}$ that is $\lipschitzconstant$ bi-\lipschitz, the corresponding inverse function $\func{\inverse{f}}{\range{f}}{\dom{f}}$ is also bi-\lipschitz with parameter $\lipschitzconstant.$
\end{lemma}

The following corollary is directly implied from Lemma~\ref{Lemma: Inverse of BiLipschitz function is BiLipschitz}.

\begin{corollary}[Bi-\lipschitz CDF implies Bi-\lipschitz Inverse CDF]\label{Corollary: BiLipschitz CDF implies BiLipschitz Inverse CDF}
	A distribution satisfying Assumption~\ref{Assumption: Bi-Lipschitz Distributions} has bi-\lipschitz inverse CDF.
\end{corollary}

\begin{definition}[$\weibull$-SubWeibull Distribution and Random Variable~\citep{VladimirovaGirardNguyenArbel/Stat/2020/SubWeibullDistributions}]
    A distribution $\wassmeas$ is said to be $\weibull$-SubWeibull if there exists some constant $\constant_\weibull$ for any $\tail \geq 0$, we have:
    \begin{align}
        \Pr_{X\sim\wassmeas}\tbrac{X \geq \tail} \leq \constant_\weibull\exp\fbrac{-\tail^{1/\weibull}}
    \end{align}
    Correspondingly, the random variable $X$ drawn from $\wassmeas$ is said to be a SubWeibull random variable.
\end{definition}

\begin{lemma}[MGF Charecterization of SubWeibull(Theorem 2.1 in~\citet{VladimirovaGirardNguyenArbel/Stat/2020/SubWeibullDistributions})]
    Given $X$ be a $\weibull$-SubWeibull random variable, then the MGF of $\abs{X}^{1/\weibull}$ satisfies:
    \begin{align*}
        \exists \constant > 0 \text{ such that } &\Exp\tbrac{\exp\fbrac{\fbrac{\gamma\abs{X}}^{1/\weibull}}}\leq \exp\fbrac{\fbrac{\gamma\constant}^{1/\weibull}}
    \end{align*}
    for all $\gamma$ such that $0 < \gamma \leq 1/\constant$.
\end{lemma}

\begin{lemma}[True Measure Concentration~\citep{InfiniteTVConcentration/Neurips20}]~\label{Lemma: Infinite TV True Measure Concentration}
For a measure $\wassmeas$ defined over a simplex $\Delta_\Nat$ and corresponding empirical distribution $\wassmeas_\wassmeassamplecount$ generated by $\wassmeassamplecount$ samples, for any $\approxerror,\confidence \in \fbrac{0,1}$, there exists a constant $\constant$ such that $\wassmeassamplecount \geq \constant\approxerror^{-2}\max\sbrac{\sum_{i \in \Nat} \sqrt{\pdf(i)},\log\fbrac{1/\confidence}}$, we have with probability $1 - \confidence$,
    \begin{align*}
        \Pr\tbrac{\TV{\wassmeas}{\empwassmeas} \geq \approxerror} \leq \confidence
    \end{align*}
\end{lemma}

\begin{lemma}[Concentration of Empirical Measure In Wasserstein Distance (Theorem 2 in~\citet{FournierGuillin/ProbandRelated/2015/ConvergenceEmpiricalWasserstein})]\label{Lemma; Concentration of Bounded-Tail Empirical Measure in Wasserstein Distance}
    Let $\wassmeas$ be a distribution on $\R$. Then for all $p \in \Nat$, 
    \begin{itemize}
        \item If $\exists \weibull < \frac{1}{p}, \gamma > 0$ such that $\Exp\tbrac{\exp\fbrac{\fbrac{\gamma\abs{X}}^{1/\weibull}}} < \infty$, then,
        \begin{align*}
            \Prob\tbrac{\Wasserstein{1}{\wassmeas}{\empwassmeas} \geq \approxerror} \leq \exp\fbrac{-\constant \wassmeassamplecount\approxerror^2} + \exp\fbrac{-\constant\wassmeassamplecount\approxerror^{1/p\weibull}}
        \end{align*}
        \item If $\exists \weibull \in \fbrac{\frac{1}{p},\infty}, \gamma > 0$ such that $\Exp\tbrac{\exp\fbrac{\fbrac{\gamma\abs{X}}^{1/\weibull}}} < \infty$, then,
        \begin{align*}
            \Prob\tbrac{\Wasserstein{1}{\wassmeas}{\empwassmeas} \geq \approxerror} \leq \exp\fbrac{-\constant \wassmeassamplecount\approxerror^2} + \exp\fbrac{-\constant\fbrac{\wassmeassamplecount\approxerror}^{1/2p\weibull}}
        \end{align*}
    \end{itemize}
\end{lemma}

The Lemma~\ref{Lemma; Concentration of Bounded-Tail Empirical Measure in Wasserstein Distance} implies the following two corollaries:

\begin{corollary}[Concentration in Wasserstein distance of Empirical Measures over $\R$~\citep{Bhat/NeurIPS/2019/ConcentrationRiskMeasuresThrough1DWasserstein}]\label{Corollary: Concentration in Wasserstein distance of 1-D Empirical Measures Subgaussian}
    Given an empirical measure $\empwassmeas$ generated by $\wassmeassamplecount$ i.i.d. samples generated from a subGaussian measure $\wassmeas$ over $\R$, we have:
    \begin{align*}
        \Prob\tbrac{\Wasserstein{1}{\wassmeas}{\empwassmeas} \geq \approxerror} \leq \exp\fbrac{-\constant \wassmeassamplecount\approxerror^2}
    \end{align*}
\end{corollary}

\begin{corollary}\label{Corollary: Concentration in Wasserstein distance of 1-D Empirical Measures Subweibull}
    Let $\wassmeas$ be a distribution on $\R$ such that $\exists \weibull \in \fbrac{1,\infty}, \gamma > 0$ such that $\Exp\tbrac{\exp\fbrac{\fbrac{\gamma\abs{X}}^{1/\weibull}}} < \infty$, then,
    \begin{align*}
        \Prob\tbrac{\Wasserstein{1}{\wassmeas}{\empwassmeas} \geq \approxerror} \leq \exp\fbrac{-\constant \wassmeassamplecount\approxerror^2} + \exp\fbrac{-\constant\fbrac{\wassmeassamplecount\approxerror}^{1/2\weibull}}
    \end{align*}
\end{corollary}

\begin{lemma}[Strong Demographic Parity and $\Wasserstein{1}{}{}$ distance~\citep{pmlr-v115-jiang20a}]\label{Lemma: Fairness quantification through Wasserstein}
	Let $\func{f}{\R^d\times[k]}{[0,1]}$ be a function where $[k]$ denotes the sensitive attribute. Let $\measure_s$ denote the output distribution of $f$ corresponding to the sensitive attribute $s \in [k]$. Then, we have for all $s,s' \in [k]$:
	\begin{align*}
		\Exp_{t \sim \uniform{[0,1]}}\tbrac{\Prob\tbrac{f(X,s) \geq t} - \Prob\tbrac{f(X,s') \geq t}} = \Wasserstein{1}{\measure_s}{\measure_{s'}}
	\end{align*}
\end{lemma}
\begin{lemma}[Privacy and Hockey Stick Divergence\citep{Balle/2018/DivergencetoPrivacy}]\label{Lemma: Privacy to Hockey Stick Divergence}
	For a given $\dpepsilon \in \R$, a mechanism $\mechanism$ is \dppair-differentially private if for all $\covariatematrix,\covariatematrix'$ with $\hamming{\covariatematrix}{\covariatematrix'} = 1$:
	\begin{align*}
		\hockeystick{e^\dpepsilon}{\mechanism\fbrac{\covariatematrix}}{\mechanism\fbrac{\covariatematrix'}} \leq \dpdelta\,
	\end{align*}
    where $\hockeystick{e^\dpepsilon}{\mu}{\nu}$ is called the Hockeystick divergence between $\mu$ and $\nu$.
\end{lemma}

\begin{lemma}[From HSD to TV Approximation~\citep{koskela2024auditingdifferentialprivacyguarantees}]\label{Lemma: Hockey Stick to TV Approxq}
    Given two measured $\wassmeas,\wassmeastwo$ and their TV distance approximations $\estwassmeas,\estwassmeastwo$ satisfying $\TV{\wassmeas}{\estwassmeas} \leq \approxerror$, and $\TV{\estwassmeas}{\estwassmeastwo} \leq \approxerror$; we have for all $\dpepsilon \in \R$:
    \begin{align*}
        \hockeystick{e^\dpepsilon}{\wassmeas}{\wassmeastwo} \leq \hockeystick{e^\dpepsilon}{\estwassmeas}{\estwassmeastwo} + \fbrac{1+e^\dpepsilon}\approxerror
    \end{align*}
\end{lemma}


%% file: Sections_PODS/Appendix/X_Experiments.tex
\newpage
\section{Experimental Details - Fairness and Privacy Auditing}\label{app:Experiments}

\subsection{Fairness Auditing} 

Group fairness measures in ML model's prediction measure the disparity in predictions of ML models across different subpopulations. The subpopulations correspond to a sensitive attribute (e.g. gender, economic status, ethnicity etc.) on which they should not be discriminated. Fairness auditors try to estimate violations of such group fairness measures by querying a model and then collecting the responses~\citep{DBLP:conf/aaai/2021/GhoshBM21/Justicia,activeaudit,AuditingFairnessbyBetting,ajarra2025active}.

A popular group fairness measure is demographic parity~\citep{feldman2015certifying}. 

\begin{definition}[Demographic Parity]\label{Definition: Demographic Parity}
	Given $O\subseteq \R$ and sensitive attributes $[k]$, a regression function $\func{f}{\R^d\times[k]}{O}$ satisfies demographic parity if for all $s,s' \in [k]$, we have	$\sup_{t \in O} \Prob(f(X,s) \leq t) - \Prob(f(X,s') \leq t) = 0$. 
    For classification, $S=[0,1]$ and the definition is referred as strong demographic parity~\citep{pmlr-v115-jiang20a}. 
\end{definition}

\cite{pmlr-v115-jiang20a} show that bounding strong demographic parity is equivalent parity is equivalent to bounding the maximum of $1$-Wasserstein distance between the output distributions of any two subpopulations. This motivates us to use \slwassbase{} to estimate demographic parity for ML models.


\textbf{Experimental Setup:}  We test accuracy and sublinearity of \slwassbase{} for fairness auditing on the well-known ACS Income dataset~\citep{Ding/NeurIPS/2021/RetiringAdult}. We test both on linear regression ($d=10$) for income as output and classification with logistic regression ($d=10$) for income above and below $40000$ USD as outputs. We compute Wasserstein Distance between the distribution of outputs of linear and logistic regression models for male and female data points of a model trained on \acsincome data. We use $3:1$ train-test split for both cases. We use scikit-learn~\citep{scikit-learn} to train both the models, and use the 'liblinear' solver for logistic regression. For reference, we compute the distance between the bucketed versions of the output distribution exactly. We choose the bucket size to be 10 and 0.01, for the regression and classification tasks, respectively. For the regression task, we increase the number of buckets as $\{500,1000,\ldots,20000\}$; and for the classification task, we increase the number of buckets as $\{50,100,\ldots,1250\}$. Finally we report the multiplicative approximation error of our estimates w.r.t. the true distance in both the cases with increasing number of buckets. We run each of the experiments $50$ times.

\textbf{Results: Sublinearity of \slwassbase{} in Fairness Auditing:} We use $416625$ total samples consisting of almost half male and female samples each.  The sample streams arrive via $\clients = 10$ sources. For the linear regression model, the approximation error drops below $0.1$ when we use $12500$ buckets. For the logistic regression model, the approximation error drops below $0.1$ when we use $750$ buckets. The difference in \#buckets is due to the difference in variance and the width of buckets used in each case.

\subsection{Privacy Auditing}



Differential privacy~\citep{Dwork2006_DP-survey} is now considered as the gold standard for data privacy protection. It aims to keep an input datum indistinguishable while looking into outputs of an algorithm.

\begin{definition}[\textbf{$(\dpepsilon,\dpdelta)$- Differential Privacy}~\citep{Dwork2006_DP-survey}]\label{Definition: Epsilon Delta DP}
	An algorithm $\func{f}{\covariatedomain}{\targetrange}$ is $(\dpepsilon,\dpdelta)$-differentially private if for any $\covariatematrix,\covariatematrix'$ with $\hamming{\covariatematrix}{\covariatematrix'} = 1$ and $\forall S \subseteq \targetrange$, we have $\Prob[f(\covariate) \in S] \leq e^{\dpepsilon} \Prob[f(\covariate') \in S] + \dpdelta$.
\end{definition}\vspace*{-.5em}

An equivalent representation of differential privacy is $		\hockeystick{e^\dpepsilon}{\mechanism\fbrac{\covariatematrix}}{\mechanism\fbrac{\covariatematrix'}} \leq \dpdelta$~\citep{Balle/2018/DivergencetoPrivacy}, where Hockey Stick Divergence (HSD) is defined as $\hockeystick{\tau}{\wassmeas}{\wassmeastwo} \triangleq \int_{\covariatedomain} [\mu(x) - \tau \nu(x)]_+ \mathrm{d}x$. 
Recently, \cite{koskela2024auditingdifferentialprivacyguarantees} show that estimating the HSD of two distributions is equivalent to estimating the HSD and TV distance between their empirical counterparts. They construct histograms over outputs of a black-box auditor and use this result to estimate TV distances for privacy auditing. As auditing privacy is data intensive, it motivates us to use \slTVbase{} in this setting. 

We adopt the experimental setting of~\citet{annamalai2024nearly} and compute the TV distance between the output distributions of logistic regressors trained on neighbouring datasets, say IN and OUT, sampled from MNIST~\citep{lecun1998mnist}. 

\textbf{Experimental Setup:} We study the performance of \slTVbase{} in the case of privacy auditing. We generate losses for datasets with and without canary using the work of~\citet{annamalai2024nearly}. We run logistic regression on \texttt{MNIST} dataset with $50$ epochs and $\dpepsilon = 10$. We denote the losses with and without canary to be \texttt{lossesin} and \texttt{lossesin}, respectively. We obtain $1000$ samples of losses for both the cases and take these losses as the distribution of interest. We choose the bucket size to be $0.01$ for the logistic regression task. The mean of \texttt{lossesin} and \texttt{lossesout} are $-2.3194$ and $-2.3241$, respectively. The standard deviation of \texttt{lossesin} and \texttt{lossesout} are $0.005654$ and $0.005535$, respectively. We increase the number of buckets as $\{10,20,\ldots,100\}$. Finally we report the multiplicative approximation error of our estimates w.r.t. the true distance in both the cases with increasing number of buckets. We run each of the experiments $50$ times.

\textbf{Results:} Given $1000$ samples, Figure~\ref{Fig: Noisy Regression} show that \slTVbase{} computes the TV distance between output distributions for IN and OUT datasets.  The sample streams arrive via $\clients = 10$ sources. The approximation error drops below $0.1$ while using only $250$ buckets. This shows \slTVbase{} can conduct resource-efficient privacy auditing of large-scale datasets.

%% file: Sections_PODS/Appendix/8_Large_Figures.tex
\newpage
\subsection{Enlarged Plots}\label{Appendix: Figures}


\begin{figure*}[ht!]
\centering
\begin{minipage}{0.48\textwidth}
\includegraphics[width=\linewidth]{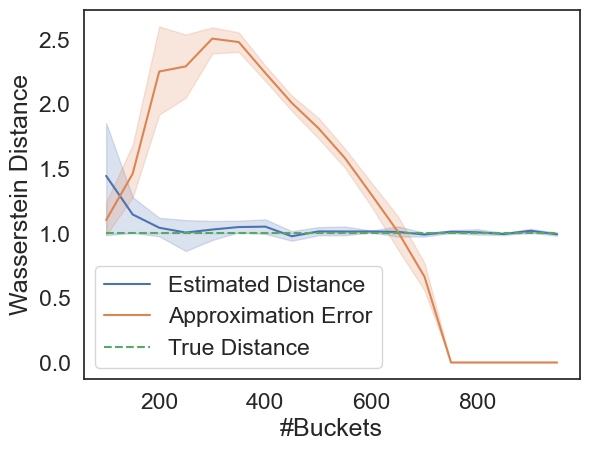}\vspace*{-1.5em}
\caption{Performance of \slwassbase{} with $\normal(0,5)$, $\normal(1,5)$ and $\bucketsize = 0.05$}
\end{minipage}
\begin{minipage}{0.48\textwidth}
\includegraphics[width=\linewidth]{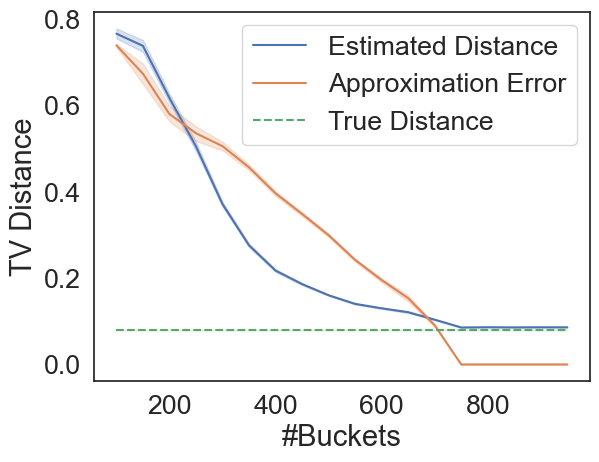}\vspace*{-1.5em}
\caption{Performance of \slTVbase{} with $\normal(0,5)$, $\normal(1,5)$ and $\bucketsize = 0.05$}
\end{minipage}
\begin{minipage}{0.48\textwidth}
\includegraphics[width=\linewidth]{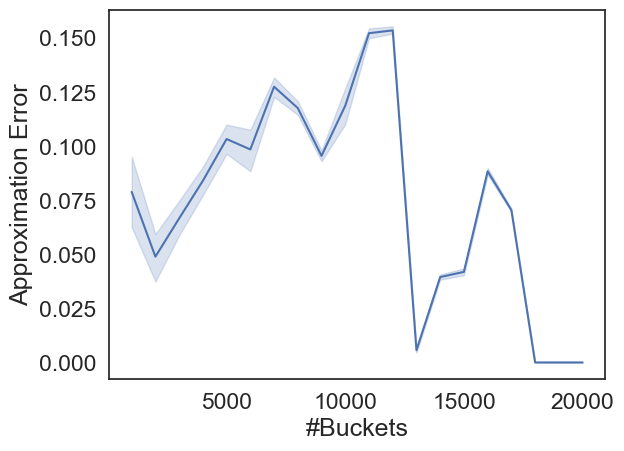}\vspace*{-1.5em}
\caption{Auditing with \slwassbase{} on regression output of \acsincome}
\end{minipage}
\begin{minipage}{0.48\textwidth}
\includegraphics[width=\linewidth]{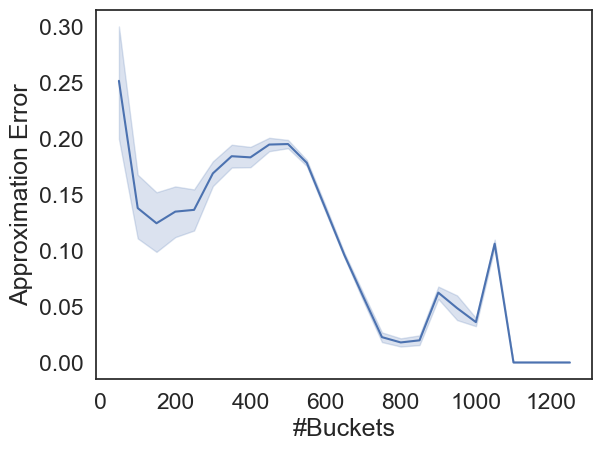}\vspace*{-1.5em}
\caption{Auditing with \slwassbase{} on classification output of \acsincome}
\end{minipage}
\begin{minipage}{0.48\textwidth}
\includegraphics[width=\linewidth]{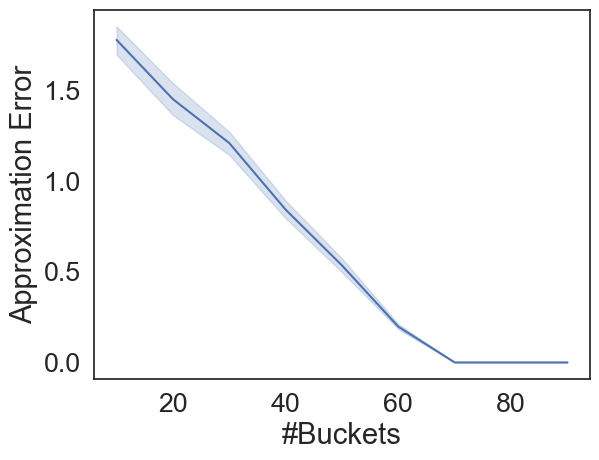}\vspace*{-1.5em}
\caption{Privacy Auditing of logistic regression on MNIST.}
\end{minipage}\vspace*{-1.5em}
\end{figure*}